\documentclass[journal]{IEEEtran}
\usepackage{tikz}
\usepackage{cite}
\usetikzlibrary{calc}
\usetikzlibrary{arrows,shapes,chains}
\usepackage{xcolor}
\usepackage{amsfonts}
\usepackage{amsmath}
\usepackage{graphicx}
\usepackage{multirow}
\usepackage{algorithmic}
\usepackage{algorithm}
\usepackage{subfigure}

\usepackage{enumitem}
\usepackage{epstopdf}
\usepackage{url}
\usepackage{bm}
\usepackage{hyperref}
\usepackage{cleveref}
\crefname{section}{Section}{Sections}
\usepackage{amsthm,amsmath,amssymb}
\usepackage{mathrsfs}
\usepackage{diagbox}
\usepackage{slashbox}
\usepackage{booktabs}
\crefname{equation}{}{}
\crefname{table}{TABLE}{TABLES}
\crefname{figure}{Fig.}{Figs.}

\newcommand{\cb}[1]{\ifmmode {\boldsymbol{#1}}\else ${\boldsymbol{#1}}$\fi}

\onecolumn
\title{DTU-Net: A Multi-Scale Dilated Transformer Network for Nonlinear Hyperspectral Unmixing}
\author{Chentong~Wang, Jincheng~Gao, Fei~Zhu, \IEEEmembership{Member,~IEEE}, Abderrahim Halimi, \IEEEmembership{Senior~Member,~IEEE} \\and C\'edric Richard, \IEEEmembership{Senior~Member,~IEEE}
  \thanks{The work was supported by the National Natural Science Foundation of China under Grant 61701337 and the UK Royal Academy of Engineering under the Research Fellowship Scheme (RF/201718/17128).~\emph{(Corresponding author: Fei Zhu)}}
	\thanks{C. Wang, J.~Gao, and F.~Zhu are with the Center for Applied Mathematics, Tianjin University, Tianjin, 300072, China (e-mail:~chentong\_wang@tju.edu.cn;2023233003@tju.edu.cn;~fei.zhu@tju.edu.cn).}
	\thanks{A. Halimi is with  the School of Engineering and Physical Sciences, Heriot-Watt University,
Edinburgh, EH14 4AS, United Kingdom (e-mail:~a.halimi@hw.ac.uk).}
	\thanks{C. Richard is with Universit\'e C\^ote d'Azur, CNRS, OCA, F-06108, Nice, France (e-mail:~cedric.richard@unice.fr).}
}

\begin{document}
\maketitle
\begin{abstract} 
Transformers have shown significant success in hyperspectral unmixing (HU). However, challenges remain.
While multi-scale and long-range spatial correlations are essential in unmixing tasks, current Transformer-based unmixing networks, built on Vision Transformer (ViT) or Swin-Transformer, struggle to capture them effectively.
Additionally, current Transformer-based unmixing networks rely on the linear mixing model, which lacks the flexibility to accommodate scenarios where nonlinear effects are significant.
To address these limitations, we propose a multi-scale Dilated Transformer-based unmixing network for nonlinear HU (DTU-Net).
The encoder employs two branches. The first one performs multi-scale spatial feature extraction using Multi-Scale Dilated Attention (MSDA) in the Dilated Transformer, which varies dilation rates across attention heads to capture long-range and multi-scale spatial correlations. The second one performs spectral feature extraction utilizing 3D-CNNs with channel attention. The outputs from both branches are then fused to integrate multi-scale spatial and spectral information, which is subsequently transformed to estimate the abundances.
The decoder is designed to accommodate both linear and nonlinear mixing scenarios. 
Its interpretability is enhanced by explicitly modeling the relationships between endmembers, abundances, and nonlinear coefficients in accordance with the polynomial post-nonlinear mixing model (PPNMM). Experiments on synthetic and real datasets validate the effectiveness of the proposed DTU-Net compared to PPNMM-derived methods and several advanced unmixing networks.
\end{abstract}
\pagebreak

\section{Introduction}
\IEEEPARstart{H}{yperspectral} imaging provides rich spectral information by capturing the land cover of interest across numerous contiguous wavelength bands, enabling applications in various fields such as classification and anomaly detection~\cite{R1}.
Despite its spectral richness, hyperspectral images often suffer from limited spatial resolution, where each pixel typically represents a mixture of multiple ground cover materials. The distinct spectral signatures associated with these materials are termed endmembers, while their proportional contributions within a pixel are referred to as abundances.
Spectral unmixing aims to identify these endmembers and estimate their corresponding abundances for each pixel~\cite{R3}. 

Before the unmixing process, a mixing model is typically assumed to describe the interaction between light and materials in hyperspectral imaging. The linear mixing model (LMM) is the most widely adopted due to its simplicity and interpretability. The LMM assumes that photons interact with a single endmember before reaching the sensor, representing each pixel as a linear combination of endmembers weighted by their respective abundances~\cite{R4}. However, the LMM often struggles in scenarios where nonlinear interactions among endmembers play a significant role.
To address this limitation, various nonlinear mixing models have been proposed to explicitly or implicitly characterize photon propagation and endmember interactions. 
Notable examples include Hapke's model for intimate mixing~\cite{hapke1981bidirectional}, kernel-based models~\cite{chen2012nonlinear, 7448928}, mismodeling-based models~\cite{sup1, sup2, sup3, sup4}, bilinear mixing models (BMMs)~\cite{nascimento2009nonlinear, R5, R6, R7}, as well as multilinear mixing models (MLMs)~\cite{R9, QiWei2017, yangbin2018} considering up to infinite-order interactions between endmembers.

Bilinear mixing, a fundamental form of nonlinear mixing, occurs when photons successively interact with two distinct endmembers before reaching the sensor. This phenomenon is particularly significant in hyperspectral images of forested regions, where interactions between soil and vegetation are prominent~\cite{R1}. 
BMMs typically extend the LMM by incorporating second-order terms to capture these endmember interactions. Mathematically, these models are expressed as the sum of an LMM component and the second-order interaction terms. Depending on the constraints applied to the interaction coefficients, various bilinear models have been proposed~\cite{nascimento2009nonlinear, R5, R6, R7}. 
While many BMMs~\cite{nascimento2009nonlinear, R5, R6} require multiple parameters to describe second-order nonlinearities at pixel level, the polynomial post nonlinear mixing model (PPNMM)~\cite{R7} offers a more compact representation by characterizing polynomial nonlinearity with a single scalar parameter $b$. Notably, it is a flexible generalization of the LMM, reverting to the LMM when $b$ equals 0. 
Although higher-order mixing models, such as Hapke's model~\cite{hapke1981bidirectional} and MLMs~\cite{R9, QiWei2017, yangbin2018} theoretically extend the modeling capabilities, the spectral amplitudes of the higher-order interactions are typically very low in practice and often indistinguishable from noise~\cite{R7}. 
To balance modeling accuracy and computational complexity, this paper adopts the PPNMM as the primary mixing model. 

Recent advancements in deep learning have been increasingly applied to unmixing, with autoencoder (AE) as a prominent framework \cite{R10}. Early AEs typically operate under the LMM and perform single-pixel processing. In general, the encoder maps observed pixels to low-dimensional representations, interpreted as abundances, while a fully connected decoder reconstructs the pixels, with its weights corresponding to the estimated endmember matrix \cite{R11}. For example, a stack of AEs detects outliers before unmixing in a final AE \cite{R12}. In \cite{R13}, uDAS is proposed to decouple the encoder and decoder, applying nonnegativity constraints to the decoder and incorporating denoising and $\ell_{21}$-norm constraints to improve estimations of both endmembers and abundances.

To better model the spectral and spatial information inherent in hyperspectral imagery, advanced deep learning techniques, examplified by convolutional neural networks (CNNs) and Transformers, have been integrated into AE-based unmixing frameworks. These techniques primarily enhance the encoder structure, enabling implicit regularization of abundance estimations. CyCU-Net \cite{R17} uses 2D convolution within a cascade autoencoder to preserve spatial details more effectively during image reconstruction. Similarly, CNNAEU \cite{R16} processes image patches, allowing neighboring pixels to contribute to the reconstruction of the central pixel. CNNs have also been incorporated into nonlinear unmixing networks \cite{R18, R19, R20, R21}, which further improve the modeling of spectral-spatial relationships. However, despite their effectiveness at extracting local spatial features, CNNs struggle to capture long-range spatial dependencies within hyperspectral images.

Transformers, initially developed for natural language processing (NLP), have achieved significant success in a wide range of domains \cite{R22}. A prominent variant, the Vision Transformer (ViT) \cite{R24}, is particularly effective at modeling long-range dependencies within images, enabling the capture of global context and relationships between distant regions. 
In the context of unmixing, DeepTrans \cite{R25} combines a CNN and ViT encoder to extract both local and global spatial features, while its CNN-based decoder is enhanced with a dual-channel graph regularizer to model semantic and manifold structural information \cite{R26}.
Despite its strengths in capturing global spatial features, ViT faces challenges in computational efficiency and struggles to fully exploit the multi-scale spatial information that is critical for hyperspectral unmixing, particularly in complex scenes with diverse target scales.
To this end, recent works \cite{R28, R29, R30} have incorporated the shifted window multi-head self-attention (SW-MSA) mechanism from the Swin Transformer \cite{9710580}. SW-MSA performs self-attention within fixed window sizes, with a window-shifting strategy in subsequent layers to enable information exchange across non-adjacent patches. For instance, UST-Net \cite{R29} integrates SW-MSA with downsampling and upsampling operations to capture multi-scale spatial features, while Swin-HU \cite{R28} applies the same mechanism to effectively model multi-scale spatial dependencies. Additionally, \cite{R30} combines a Transformer encoder with a CNN encoder, using a window-based pixel-level multi-head self-attention (WP-MSA) mechanism to capture non-local features and preserve spatial resolution by leveraging pixel embeddings, enhancing spatial fidelity in sparse unmixing.

However, using SW-MSA to model multi-scale spatial information in unmixing has certain limitations. 
This mechanism primarily emphasizes local self-attention, capturing interactions between patches within small receptive fields, which restricts its capacity to capture long-range dependencies across the entire hyperspectral image.
Moreover, the choice of window size is case-specific: a small window size limits the receptive field, preventing the capture of global context, while a large window results in redundant attention computations, increasing computational costs. 

While many existing unmixing networks are based on LMM, others have been adapted to perform nonlinear unmixing by incorporating well-established nonlinear unmixing models in their decoder designs \cite{R14}. 
For example, HapkeCNN \cite{9869671} is specifically designed for intimate mixtures \cite{hapke1981bidirectional}, while 3D-NAE \cite{R18} is tailored to the linear-mixture/nonlinear-fluctuation mixing model \cite{chen2012nonlinear}. Additionally, recent MLM-driven unmixing networks \cite{R21,2024EMLM} have been introduced to enhance the physical interpretability of the network structure.

As a key category of nonlinear models, BMMs have been employed as the primary mixing model for network design in several works.
For example, \cite{R19} incorporates the GBM within the decoder to extract endmembers and estimate second-order scattering interactions, with secondary interaction coefficients treated as trainable network parameters. Based on the PPNMM, \cite{2019Nonlinear} uses an implicit mapping to approximate the post-nonlinear term. However, this method deviates from the original formulation, thus lacking interpretability, and cannot estimate pixel-level nonlinear coefficients.
In contrast, \cite{R14} introduces PPNMM-AE, which explicitly models the nonlinear coefficients of PPNMM as learnable parameters. Similarly, \cite{R20} integrates a PPNMM-based decoder with superpixel-based graph convolutional networks to capture global spatial information, with nonlinear coefficients also learned as network parameters. 
However, treating pixel-wise nonlinear coefficients as network parameters, rather than as features associated with input pixels (similar to abundance), restricts the model adaptability in adjusting the size and shape of parameters for different input datasets.

\subsection*{Motivations and Contributions}

In unmixing tasks, accurate abundance estimation necessitates effective modeling of both global and multi-scale spatial information inherent in hyperspectral images. Although current Transformer-based approaches that utilize the SW-MSA mechanism from the Swin Transformer are effective in alleviating the computational burden of ViT, they struggle to capture multi-scale and long-range dependencies due to the localized nature of SW-MSA. This limitation becomes particularly problematic while the global and multi-scale context is essential for hyperspectral images unmixing. Therefore, advanced attention mechanisms that extend beyond SW-MSA from the Swin Transformer are required to better capture semantic dependencies across multiple scales in unmixing.

Furthermore, current Transformer-based unmixing networks~\cite{R25,R26,R28,R29,R30} only rely on LMMs and lack compatibility with nonlinear models, leading to degraded performance when the mixing/unmixing models are mismatched. 
Specifically, current unmixing networks based on BMMs suffer from limited interpretability and flexibility.
These networks either rely on implicit mappings to model nonlinear terms \cite{R18,2019Nonlinear} or explicitly treat pixel-level nonlinear coefficients as network parameters \cite{R14,R19,R20}, rather than learning them as features directly from the input data.

To address the aforementioned limitations, this paper presents an end-to-end multi-scale Dilated Transformer-based nonlinear Unmixing Network (DTU-Net). In the encoder, we utilize Dilated Transformer to capture multi-scale spatial dependencies and incorporate spectral dependencies through 3D-CNNs and channel attention, effectively fusing spatial and spectral features. The decoder is designed based on the PPNMM mechanism, enabling explicit nonlinear unmixing and ensuring network interpretability.  
The main contributions are summarized as follows:

\begin{enumerate}
\item We introduce a novel encoder branch based on Dilated Transformer \cite{R27}, which captures multi-scale spatial dependencies in hyperspectral images. This branch applies Sliding Window Dilated Attention (SWDA) to compute sparse self-attention across selected patches. We then apply Multi-Scale Dilated Attention (MSDA) block, which varies dilation rates across attention heads to enable multi-scale representation, capturing both long-range and multi-scale spatial correlations in hyperspectral images.

\item We incorporate 3D-CNNs and channel attention from CBAM \cite{R31} to construct a second encoder branch, which mainly captures spectral dependencies in hyperspectral images. 
The output features from both branches are fused, allowing the encoder to fully leverage both spatial and spectral information.

\item We propose an unsupervised nonlinear unmixing network based on the PPNMM, which enables flexible modeling of both linear and nonlinear mixing scenarios. 
The explicit unmixing is realized primarily in the decoder. To enhance interpretability, the decoder leverages learned features and specialized layers to model the relationships between endmembers, abundances, and nonlinear coefficients in accordance with the PPNMM. Instead of treating nonlinear coefficients as network parameters or modeling them implicitly by network layers, we learn them as pixel-wise features, improving the network's adaptability across diverse datasets.\end{enumerate}

The rest of the paper is organized as follows. Section \ref{Sec:II} reviews the relevant unmixing model and attention mechanisms. Section \ref{Sec:III} presents the multi-scale Dilate Transformer network for nonlinear unmixing. Experimental results are presented in Section \ref{Sec:IV}, and Section \ref{Sec:V} concludes the paper. 

\section{Related Works}\label{Sec:II}
We begin by briefly reviewing the mixing model used in this paper, followed by an introduction to the multi-scale dilate attention mechanism.
\subsection{Polynomial Post Nonlinear Mixing Model (PPNMM)}
\label{ppnmm}
The PPNMM is a widely-used model for nonlinear unmixing, capturing up to second-order interactions between endmembers \cite{R7}. 
Let $\boldsymbol{y} \in \mathbb{R}^{L \times 1}$ represent the observed spectrum of a pixel across $L$ spectral bands.
The PPNMM is expressed as a linear mixing term augmented by a polynomial nonlinear term, which is given by
\begin{equation}
	\label{PPNMM}
	\boldsymbol{y}=\boldsymbol{M}^{\top}\boldsymbol{a}+b (\boldsymbol{M}^{\top}\boldsymbol{a})\odot (\boldsymbol{M}^{\top}\boldsymbol{a})+\boldsymbol{n},
\end{equation}
where $\odot$ denotes the Hadamard product, $\boldsymbol{M} = [\boldsymbol{m}_1, \boldsymbol{m}_2, \dots, \boldsymbol{m}_R]^{\top} \in \mathbb{R}^{R \times L}$ is the endmember matrix composed by $R$ endmembers,  $\boldsymbol{a} = [a_1, a_2, \dots, a_R]^\top \in \mathbb{R}^{R \times 1}$ is the abundance vector, $b$ is a coefficient that balances linear and bilinear effects, and $\boldsymbol{n} \in \mathbb{R}^{L \times 1}$ represents the additive Gaussian noise.
The original PPNMM \cite{R7} is a supervised unmixing method, where the endmember matrix $\boldsymbol{M}$ is known in prior. 
Due to physical interpretation, both the abundance non-negative constraint (ANC) and sum-to-one constraint (ASC) constraints are imposed, with $a_{r}\ge 0~\text{and}~\sum_{r=1}^{R}a_{r}=1$.

According to \eqref{PPNMM}, the pixel-wise nonlinear effects are characterized by a single scalar parameter $b$, resulting in a simpler model compared to other second-order nonlinear unmixing models \cite{R5}\cite{R6}. Notably, when $b = 0$, PPNMM reduces to LMM, which is given by 
\begin{equation}
	\label{LMM}
	\boldsymbol{y}=\boldsymbol{M}^{\top}\boldsymbol{a}+\boldsymbol{n}.
\end{equation}
As a result, the PPNMM is a flexible model to model both linear and nonlinear mixing scenarios.

\subsection{Multi-Scale Dilated Attention (MSDA)}
The Multi-Scale Dilated Attention (MSDA), proposed in DilateFormer \cite{R27}, is designed to efficiently capture multi-scale semantic information while reducing redundancy in the attention mechanism. Unlike traditional self-attention mechanisms which focus on locality in shallow layers, MSDA incorporates both locality and sparsity to aggregate information from different scales without introducing significant computation overhead.
\subsubsection{Sliding Window Dilated Attention (SWDA)}
Analogous to the self-attention mechanism in Transformers \cite{R22, R24}, the core of MSDA is based on the Sliding Window Dilated Attention (SWDA), where keys and values are  {sparsely} selected in a sliding window surrounding the query, followed by self-attention performed on these selected key/value pairs.
Let $\boldsymbol{Q},\boldsymbol{K},\boldsymbol{V}$ represent the query, key, and value matrices, respectively, where each row is a feature vector of dimension $d$. 
The output feature map $\boldsymbol{X}$ from SWDA is given by
\begin{equation}
	\boldsymbol{X}= {\rm SWDA}(\boldsymbol{Q}, \boldsymbol{K}, \boldsymbol{V}, r),
	\end{equation}
where $r$ is the dilation rate controlling the sparsity of the attention mechanism.
For a given location $(i,j)$ in $\boldsymbol{X}$, the value is computed as
\begin{equation}
	x_{ij}={\rm Softmax}(\frac {q_{ij}\boldsymbol{k}_{r}^{\top}}{\sqrt{d}})\boldsymbol{v}_{r},
\end{equation}
where $q_{ij}$ is the corresponding element in $\boldsymbol{Q}$, and  
$\boldsymbol{k}_{r}$ and $\boldsymbol{v}_{r}$ are the key and value vectors sparsely selected from $\boldsymbol{K}$ and $\boldsymbol{V}$, respectively, according to the dilation rate $r$.
The selection of $\boldsymbol{k}_{r}$ and $\boldsymbol{v}_{r}$ is based on the following coordinate system
\begin{equation}
	\begin{aligned}
         \Big\{(i',j')\Big|i'&=i+p\times r,j'=j+q\times r\Big\},\\&-\frac{w}{2}\leq p, q\leq\frac{w}{2}
	\end{aligned}
\end{equation}
where $w$ is the window size centered at coordinate $(i,j)$, and $(p,q)$ are the relative shifts within the window defined by the dilation rate $r$. 
\subsubsection{Multi-Scale Dilated Attention (MSDA) Block}
\begin{figure}[htbp]
	\centering
	\graphicspath{{Pic/}}
\subfigure[]{
		\label{MSDA}
		\includegraphics[trim=7mm 4mm 9mm 7mm, clip,width=0.41 \textwidth]{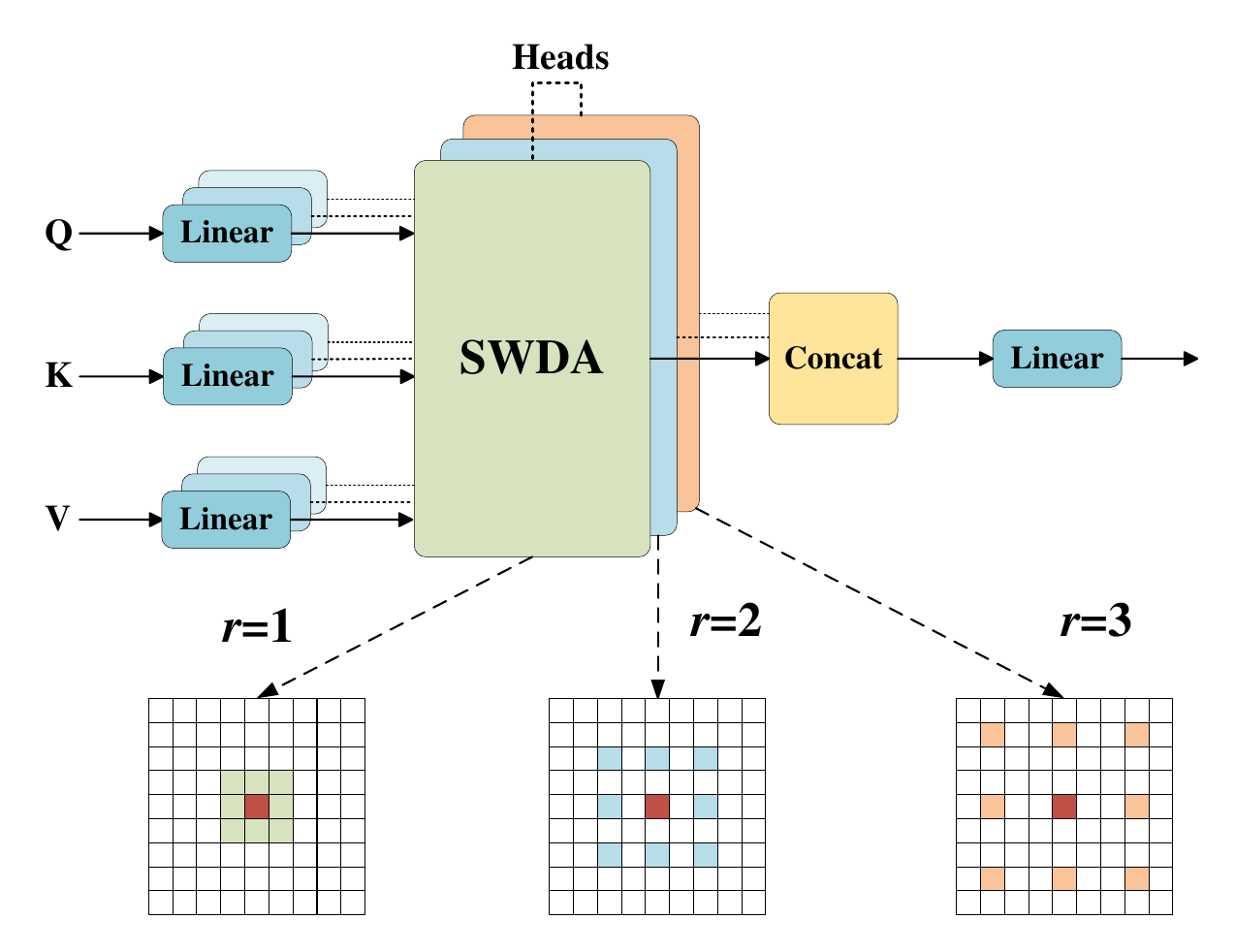} 
	}   
\subfigure[]{
		\label{MSDA_Block}
		\includegraphics[trim=14mm 15mm 13mm 10mm, clip,width=0.41 \textwidth]{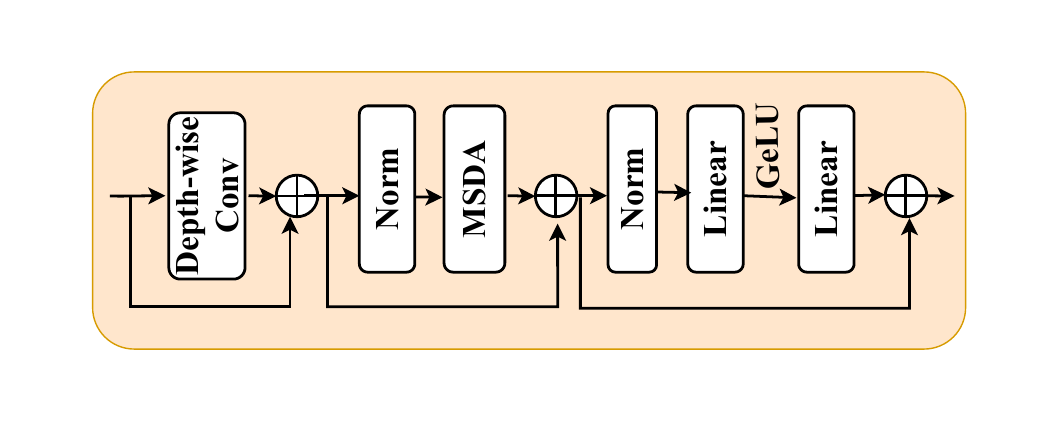} 
	}   
\caption{Illustration of Multi-Scale Dilated Attention (MSDA) \cite{R27}: (a) The MSDA mechanism; (b) Structure of an MSDA block.}
	\centering 
	\label{F1}
\end{figure}

The core of the MSDA block is the MSDA mechanism, which efficiently aggregates multi-scale spatial information by leveraging different dilation rates across various attention heads. Similar to the Multi-Head Self-Attention (MHSA) mechanism in traditional Transformers \cite{R22, R24}, the feature map is divided into multiple heads along the channel dimension before computing multi-scale attention. For each head, SWDA is computed with varying dilation rates. This mechanism is illustrated in Fig.~\ref{MSDA} and is formulated as
\begin{equation}
   \boldsymbol{h}_{i}={\rm SWDA}(\boldsymbol{Q}_{i}, \boldsymbol{K}_{i}, \boldsymbol{V}_{i}, r_{i}), i=1,\cdots,m
\end{equation}
\begin{equation}
	\boldsymbol{\mathcal{X}}={\rm Linear}({\rm Concat}[\boldsymbol{h}_{1},\cdots ,\boldsymbol{h}_{m}])
\end{equation}
where $m$ is the number of heads, $\boldsymbol{Q}_{i}$, $\boldsymbol{K}_{i}$ and $\boldsymbol{V}_{i}$ represent the query, key, and value slices of the feature map for the $i$-th head, respectively, $\boldsymbol{h}_{i}$ is the output of $i$-th head, and $\boldsymbol{\mathcal{X}}$ is the output tensor that integrates attention information from multiple scales. 

The structure of an MSDA block is illustrated Fig.~\ref{MSDA_Block}.
It begins with a depth-wise convolution with zero-padding and a kernel size of $3 \times 3$, which applies Conditional Position Embedding (CPE) to the feature map before the MSDA. The multi-scale spatial dependencies are then captured by the MSDA mechanism. Subsequently, the output traverses through a linear layer, followed by Gaussian Error Linear Unit (GeLU) activation, and another linear layer. Residual connections are consistently applied throughout the network.

\section{Proposed Method}\label{Sec:III}
\label{S3}
We propose a multi-scale Dilated Transformer unmixing network (DTU-Net) for nonlinear unmixing, which is primarily based on the PPNMM mechanism described in Section~\ref{ppnmm}. In the proposed network, the full image serves as the input for unsupervised unmixing, jointly estimating the endmembers, abundances, and nonlinear coefficients for all pixels. As a result, we first extend the pixel-wise PPNMM in \eqref{PPNMM} to the tensor-based formulation that considers the entire hyperspectral image as a whole, preserving its inherent spectral-spatial structure.

Let tensor \( \boldsymbol{\mathcal{Y}} \in \mathbb{R}^{n_\text{row} \times n_\text{col} \times L} \) represent the observed hyperspectral image. 
The tensor-based formulation of PPNMM in \eqref{PPNMM} is given by
\begin{equation}\label{PPNMM-tensor}
\boldsymbol{\mathcal{Y}} = \boldsymbol{\mathcal{A}} \times_3 \boldsymbol{M} + \boldsymbol{\mathcal{B}} \odot (\boldsymbol{\mathcal{A}} \times_3 \boldsymbol{M}) \odot (\boldsymbol{\mathcal{A}} \times_3 \boldsymbol{M}) + \boldsymbol{\mathcal{N}},
\end{equation}
where \( \boldsymbol{M} \in \mathbb{R}^{R \times L} \) is the endmember matrix, \( \boldsymbol{\mathcal{A}} \in \mathbb{R}^{n_\text{row} \times n_\text{col} \times R} \), \( \boldsymbol{\mathcal{B}} \in \mathbb{R}^{n_\text{row} \times n_\text{col} \times 1} \) and \( \boldsymbol{\mathcal{N}} \in \mathbb{R}^{n_\text{row} \times n_\text{col} \times L} \) represent the abundance tensor, the nonlinear coefficient tensor and the additive noise tensor, respectively. The operation \( \times_3 \) denotes mode-3 multiplication.
Broadcasting is used to align the third dimensions of \( \boldsymbol{\mathcal{B}} \) and \( (\boldsymbol{\mathcal{A}} \times_3 \boldsymbol{M}) \), enabling element-wise multiplication across the spatial dimensions while expanding \( \boldsymbol{\mathcal{B}} \) to match the third dimension $L$.

In primary, we design the DTU-Net to solve the following optimization problem:
\begin{equation}~\label{optimizaiton2}
\begin{aligned}
\mathop{\arg \min}_{\boldsymbol{M}\geq 0, \boldsymbol{\mathcal{A}}\geq 0, \boldsymbol{\mathcal{B}}} \!&\left\| \boldsymbol{\mathcal{Y}} \! - \! \boldsymbol{\mathcal{A}} \! \times_3  \!\boldsymbol{M} \! - \!\boldsymbol{\mathcal{B}} \odot (\boldsymbol{\mathcal{A}}  \! \times_3 \! \boldsymbol{M}) \odot (\boldsymbol{\mathcal{A}} \! \times_3 \! \boldsymbol{M}) \right\|_F^2 \\
&+ \mathcal{R}(\boldsymbol{\mathcal{A}}) + \mathcal{R}(\boldsymbol{\mathcal{B}}),\\
&\text{s.t.} \sum_{r=1}^{R} \boldsymbol{\mathcal{A}}_{i,j,r} = 1, \quad \forall i \in [1, n_\text{row}], \, j \in [1, n_\text{col}],
\end{aligned}
\end{equation}
where $\| \cdot \|_F$ denotes the Frobenius norm. Both the abundance ANC and ASC constraints are enforced, and the endmembers are constrained to be nonnegative.
The terms $\mathcal{R}(\boldsymbol{\mathcal{A}})$ and $\mathcal{R}(\boldsymbol{\mathcal{B}})$ represent the regularization effects implicitly enforced by the network structures, (e.g., convolutional layers and Transformers) on the abundance and nonlinear coefficient tensors, respectively.  
The number of endmembers $R$ is assumed to be estimated a priori, as in \cite{R33, sup5}.

\subsection{Network Overview}
\begin{figure*}
	\centering
	\graphicspath{{Pic/}}
	\includegraphics[trim=5mm 30mm 5mm 20mm, clip,width=7in]{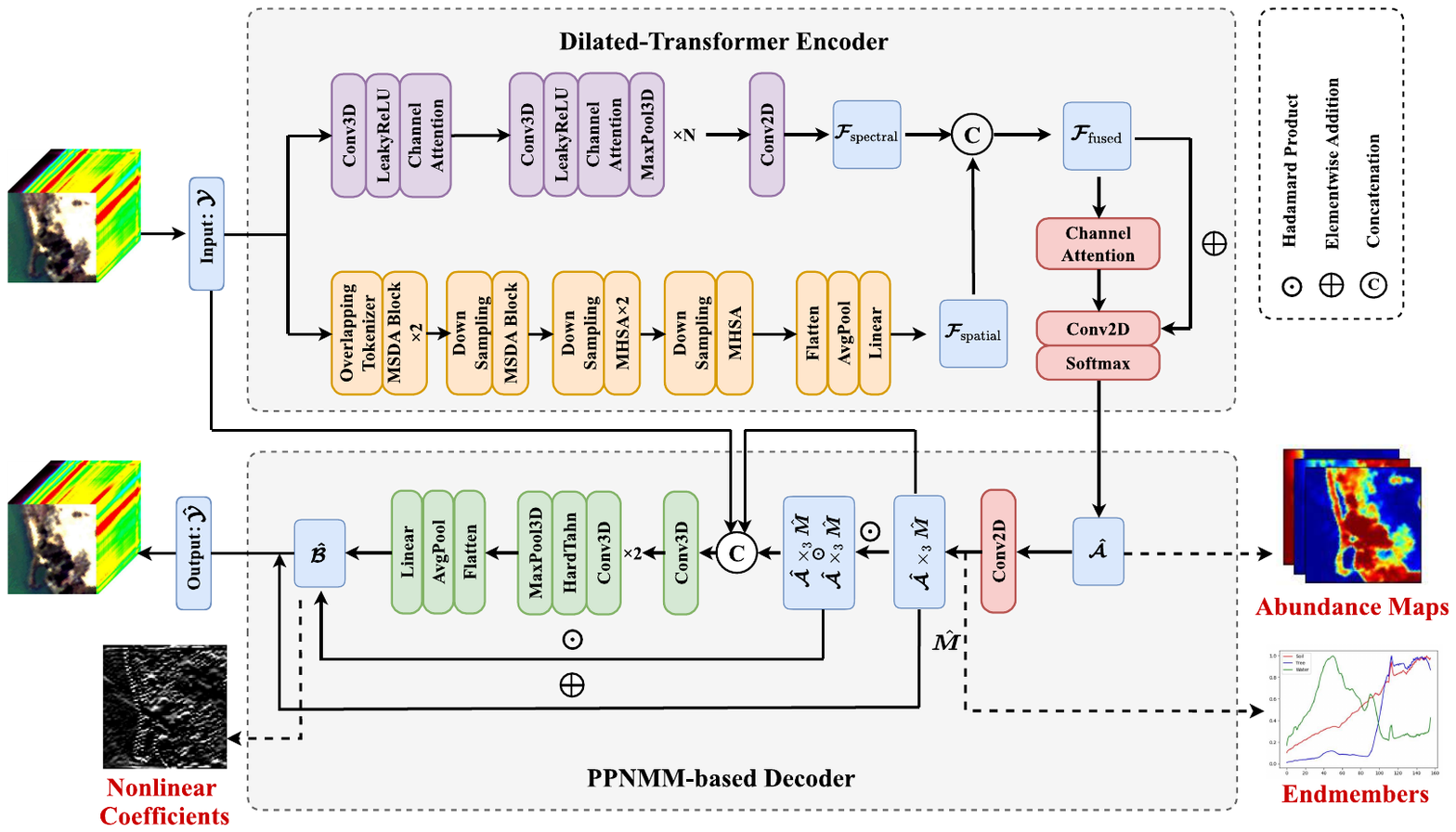}
\caption{Overview of the proposed Multi-Scale Dilated Transformer Unmixing Network (DTU-Net). DTU-Net consists of two main components: the Dilated-Transformer encoder and the PPNMM-based decoder. The encoder extracts multi-scale spatial and spectral features from the hyperspectral image through two distinct branches, which are then fused to generate the abundance tensor. The decoder applies a 2D convolutional layer to upsample the abundance tensor, with the weight matrix representing the endmember matrix, corresponding to the linear mixing part. The nonlinear coefficient tensor is learned by concatenating and transforming relevant features. Finally, the hyperspectral image is reconstructed by combining the linear and nonlinear mixing components using learned features and network parameters, following the PPNMM mechanism.}
	\label{F3} 
\end{figure*}

The architecture of the proposed DTU-Net is illustrated in Fig. \ref{F3}.
Building upon the optimization problem in \eqref{optimizaiton2}, the network consists of two main components: the Dilated Transformer encoder and the PPNMM-based decoder. The encoder includes two branches: one dedicated to multi-scale spatial feature extraction and the other to spectral feature extraction. The features from both branches are fused and subjected to dimensionality reduction to generate the estimated abundances. 
The decoder primarily employs the PPNMM mechanism to reconstruct the hyperspectral image. It first applies convolutional layer to upsample the estimated abundances, with the weights representing the endmember matrix. The nonlinear coefficients are learned as pixel-wise features.

\subsection{Dilated-Transformer Encoder}\
\textit{1) Multi-Scale Spatial Feature Extraction Branch}:
The spatial feature extraction branch follows a pyramid structure, consisting of five modules designed to extract multi-scale spatial features. As noted in \cite{R27}, shallow layers in ViTs exhibit significant sparsity and locality. To this end, two modules with Multi-Scale Dilated Attention (MSDA) blocks are applied initially, followed by two modules with traditional Multi-Head Self-Attention (MHSA) blocks.

In the first module, the overlapping tokenizer\cite{R35} which uses multiple $3 \times3$ convolution layers with zero-padding is used for patch embedding, where the channel dimension of the embedded feature map $C$ is data specific, as to be discussed in Section~\ref{subsec:parameters}. 
The multi-scale sparse attention is then computed using two MSDA blocks, each with a specified number of attention heads. For example, $\boldsymbol{\mathcal{T}}_2 = {\rm MSDA_3}(\boldsymbol{\mathcal{T}}_1)$ denotes the application of an MSDA block with 3 attention heads to $\boldsymbol{\mathcal{T}}_1$.
The operations in this module are expressed as:
\begin{equation}
	\boldsymbol{\mathcal{T}}_{1}={\rm OverlappingTokenizer}(\boldsymbol{\mathcal{Y}})
\end{equation}
\begin{equation}
	\boldsymbol{\mathcal{T}}_{2}={\rm MSDA_3}(\boldsymbol{\mathcal{T}}_{1})
\end{equation}
\begin{equation}
	\boldsymbol{\mathcal{F}}_{1}={\rm MSDA_3}(\boldsymbol{\mathcal{T}}_{2})
\end{equation}
where $\boldsymbol{\mathcal{F}}_{1}$ denotes the output of the first module.

In the second module, a convolutional layer with a $3 \times 3$ kernel and a $2 \times 2$ stride performs downsampling, integrating the local features of adjacent pixels in the feature map. After downsampling, the feature is passed through an MSDA block with 6 heads, where multi-scale attention is calculated. The operations in this module are expressed as
\begin{equation}
	\boldsymbol{\mathcal{T}}_{3}={\rm DownSampling}(\boldsymbol{\mathcal{F}}_{1})
\end{equation}
\begin{equation}
	\boldsymbol{\mathcal{F}}_{2}={\rm MSDA_6}(\boldsymbol{\mathcal{T}}_{3})
\end{equation}
where $\boldsymbol{\mathcal{F}}_{2}$ is the output of the second module.

In the third and fourth modules, the same downsampling operation was carried out. Because the size of the feature map has been reduced to $1/16$ of its original size, these two modules differ in that the MSDA block is substituted by the MHSA block. The operation of the two modules is expressed by 
\begin{equation}
	\boldsymbol{\mathcal{T}}_{4}={\rm DownSampling}(\boldsymbol{\mathcal{F}}_{2})
\end{equation}
\begin{equation}
	\boldsymbol{\mathcal{T}}_{5}={\rm MHSA_{12}}(\boldsymbol{\mathcal{T}}_{4})
\end{equation}
\begin{equation}
	\boldsymbol{\mathcal{F}}_{3}={\rm MHSA_{12}}(\boldsymbol{\mathcal{T}}_{5})
\end{equation}
\begin{equation}
	\boldsymbol{\mathcal{T}}_{6}={\rm DownSampling}(\boldsymbol{\mathcal{F}}_{3})
\end{equation}
\begin{equation}
	\boldsymbol{\mathcal{F}}_{4}={\rm MHSA_{24}}(\boldsymbol{\mathcal{T}}_{6})
\end{equation}
where $\boldsymbol{\mathcal{F}}_{3}$, $\boldsymbol{\mathcal{F}}_{4}$ denote the output of the third and forth module, respectively. The two MHSA blocks in the third module and the one MHSA block in the forth module are respectively set up with 12 heads and 24 heads.

After computing multi-scale and global attention across all four modules, the feature map effectively integrates both local contextual features and global information from the original image. The feature is then transformed into the spatial feature map $\boldsymbol{\mathcal{F}}_\text{spatial}$, which is aligned in size with the abundance tensor, through the processes of upsampling via average pooling, followed by flattening and a linear layer.

\textit{2) Spectral Feature Extraction Branch:}
In the spectral feature extraction branch, the process initiates with a convolutional layer with a kernel size of $3 \times 1 \times 1$, combined with a LeakyReLU activation function and channel attention as proposed in CBAM \cite{R31}. This setup performs initial dimensionality reduction on the input observed image $\boldsymbol{\mathcal{Y}}$. Subsequently, a series of 3D convolutional layers with $3 \times 1 \times 1$ kernel sizes, followed by LeakyReLU activation functions and 3D max-pooling layers with $2 \times 1 \times 1$ window sizes, along with channel attention mechanisms, are employed to capture the spectral features of the image. Throughout this process, the spectral dimension of the hyperspectral image $L$ is progressively reduced. Here, the number of series, denoted as $N$, is data specific.
Finally, a 2D convolutional layer with a $3 \times 3$ kernel size reduces the spectral dimension of $\boldsymbol{\mathcal{F}}_\text{spectral}$ to the number of endmembers $R$, resulting in a feature map of the same size as the abundance tensor.

\textit{3) Feature Fusion:}
To comprehensively integrate the spatial and spectral information, $\boldsymbol{\mathcal{F}}_\text{spatial}$ and $\boldsymbol{\mathcal{F}}_\text{spectral}$ are concatenated along the channel dimension, as follows:
\begin{equation} 
\boldsymbol{\mathcal{F}}_\text{fused} = {\rm Concat}[\boldsymbol{\mathcal{F}}_\text{spatial}, \boldsymbol{\mathcal{F}}_\text{spectral}], 
\end{equation}
where $\boldsymbol{\mathcal{F}}_\text{fused}$ represents the fused feature map that combines both the multi-scale spatial information and the spectral information derived from the original hyperspectral image.
This fusion ensures that complementary spatial and spectral characteristics are effectively preserved and utilized. To further refine the fused feature, channel attention is calculated to emphasize the spectral features. The resulting attention-weighted feature is then pixel-wise added to the initial fused feature for enhancement.
Next, a 2-D convolutional layer with a $3 \times 3$ kernel and zero-padding is applied to reduce the channel dimensionality, ensuring it matches the number of endmembers.
The Softmax function is finally applied to satisfy the abundance ANC and ASC constraints, as given by
\begin{equation} \label{softmax}
\boldsymbol{\mathcal{\hat{A}}} = {\rm Softmax}(\gamma \boldsymbol{\mathcal{F}})
\end{equation}
where $\boldsymbol{\mathcal{F}}$ denotes the output from the previous processing stage and $\gamma \geqslant 1$. A higher value of $\gamma$ induces a similar effect to ${\ell _{1}}$-norm regularization on the abundance maps \cite{R29,R36}, thereby enhancing sparsity.

\subsection{PPNMM-based Decoder}
The decoder of DTU-Net is designed based on the PPNMM formulation in~\eqref{PPNMM-tensor}, aiming to reconstruct the hyperspectral image while adhering to the physical principles of PPNMM. This subsection details the network design for the linear and second-order nonlinear mixing components.

\textit{1) Linear Mixing Component:}
The decoder receives the estimated abundance tensor \( \boldsymbol{\mathcal{\hat{A}}} \) as input. A \(1 \times 1\) 2D convolutional layer with a stride of \(1 \times 1\) processes \(\boldsymbol{\mathcal{\hat{A}}}\) to simulate a fully connected layer. This operation transforms the feature dimension from \(R\) to \(L\), producing an output feature map with spatial dimensions matching the original hyperspectral image. The convolution kernel size is \(1 \times 1 \times R \times L\), and the operation can be expressed as:
\begin{equation}\label{PPNMM-Linear}
\boldsymbol{\mathcal{Y}}_{\text{linear}} = \boldsymbol{{\mathcal{\hat{A}}}} \times_3 \boldsymbol{W} = \mathrm{Conv2D}(\boldsymbol{{\mathcal{\hat{A}}}}),
\end{equation}
where \(\boldsymbol{W}\) is the weight matrix of the convolution kernel, interpreted as the estimated endmember matrix \(\boldsymbol{\hat{M}}\). Specifically, the weight matrix \(\boldsymbol{W}\) is enforced to satisfy the non-negativity constraint, which is implemented by
\begin{equation}
	\phi(x)={\rm max}(0,x),
\end{equation}
where $x$ is an element of $\boldsymbol{W}$.
In this work, \(\boldsymbol{W}\) is initialized using the results from VCA~\cite{R37}.

\textit{2) Nonlinear Mixing Component:}
A major challenge in parameter estimation lies in accurately learning the nonlinear coefficients in model~\eqref{PPNMM}.
Unlike previous methods~\cite{R19,R14,R20}, which treat pixel-wise nonlinear coefficients as network parameters or use network layers (such as fully connected layers) to implicitly model nonlinear interactions among endmembers~\cite{2019Nonlinear}, we propose learning these coefficients as features, similar to abundances. This helps to improve network interpretability and stability while adhering to the physical principles of PPNMM.

As described in \eqref{optimizaiton2}, the estimation of the nonlinear coefficient tensor \(\boldsymbol{\mathcal{B}}\) is relevant to three components: the observed hyperspectral image \(\boldsymbol{\mathcal{Y}}\), the linear mixing part \(\boldsymbol{\mathcal{A}} \times_3 \boldsymbol{M}\), and the second-order interaction term \((\boldsymbol{\mathcal{A}} \times_3 \boldsymbol{M}) \odot (\boldsymbol{\mathcal{A}} \times_3 \boldsymbol{M})\). 
To this end, the corresponding tensors are concatenated along the channel dimension, forming a feature map with an expanded channel size of $3L$ that encodes rich nonlinear information.
The feature map is processed through a 3-D convolutional layer with $5\times 1\times 1$ kernel size to extract spatial-spectral patterns, followed by two modules consisting of 3-D convolutional layers of the same kernel size, HardTanh activations, and 3-D max-pooling layers with $2\times 1\times 1$ window size. These modules progressively reduce the channel dimension while preserving key features. 
Following the series of dimensionality reduction processes, the feature map is flattened, averaged, and passed through a linear layer, producing a single-channel feature map $\boldsymbol{\mathcal{\hat{B}}}$, which estimates the nonlinear coefficient tensor for the whole image.

Using the estimated endmember matrix, abundances, and nonlinear coefficients, the second-order mixing component of the hyperspectral image is modeled as
\begin{equation}\label{PPNMM-Nonlinear}
\boldsymbol{\mathcal{Y}}_{\text{nonlinear}} =  \boldsymbol{\mathcal{\hat{B}}} \odot (\boldsymbol{\mathcal{\hat{A}}} \times_3 \hat{\boldsymbol{M}}) \odot (\boldsymbol{{\mathcal{\hat{A}}}}\times_3 \hat{\boldsymbol{M}}).
\end{equation}

Finally, the reconstructed hyperspectral image $\boldsymbol{\mathcal{\hat{Y}}}$ is obtained in accordance with the model in \eqref{PPNMM-tensor}, by adding both the linear part~\eqref{PPNMM-Linear} and nonlinear mixing part \eqref{PPNMM-Nonlinear}.

\subsection{Loss Function}
To train the proposed network model, two complementary losses are utilized, namely reconstruction error (RE) loss and spectral angle distance (SAD) loss, which are defined as 
\begin{equation}
  L_{\mathrm{RE}}(\boldsymbol{\mathcal{Y}}, \boldsymbol{\mathcal{\hat{Y}}}) = \frac{1}{N} \|\boldsymbol{\mathcal{\hat{Y}}}- \boldsymbol{\mathcal{Y}}\|_{F}^{2},
\end{equation}

\begin{equation}
  L_{\mathrm{SAD}}(\boldsymbol{\mathcal{Y}}, \boldsymbol{\mathcal{\hat{Y}}}) = \frac{1}{N} \sum_{i,j} \arccos \left( \frac{\langle \boldsymbol{\mathcal{Y}}_{i,j,:}, \boldsymbol{\mathcal{\hat{Y}}}_{i,j,:} \rangle}{\|\boldsymbol{\mathcal{Y}}_{i,j,:}\|_{2} \|\boldsymbol{\mathcal{\hat{Y}}}_{i,j,:}\|_{2}} \right),
\end{equation}
where $N={n_\text{row} \times n_\text{col}} $ represents the total number of pixels of the image.
The RE loss is computed as the mean squared error between the observed and reconstructed images, quantifying the pixel-wise difference. While this loss effectively guides the model in accurately reconstructing individual pixels, it is sensitive to spectral scale variations due to the high dimensionality of hyperspectral images.
On the other hand, the SAD loss is scale-invariant and effectively recovers the spatial distribution of features. Following~\cite{R25, R29}, we define the total loss function as a weighted sum of the two losses
\begin{equation}\label{loss}
	L=\alpha L_{\mathrm{RE}}+L_{\mathrm{SAD}},
\end{equation}
where $\alpha$ is a hyperparameter that balances the two losses.

\section{Experiments}\label{Sec:IV}
In this section, experiments were conducted on a series of synthetic datasets and three real hyperspectral images, to evaluate the unmixing performance of the proposed DTU-Net. 
Seven state-of-the-art methods were selected for comparison.
These include the supervised PPNMM~\cite{R7} and six advanced unmixing networks designed for both linear and nonlinear mixtures, based on autoencoders, CNNs or Transformers.
Specifically, we considered the Cycle-Consistency Unmixing Network (CyCU-Net)~\cite{R17}, the deep transformer network (DeepTrans)~\cite{R25}, the Swin Transformer unmixing network (Swin-HU)~\cite{R28}, and the U-shaped Transformer network using shifted windows (UST-Net)~\cite{R29} proposed for linear unmixing, as well as the PPNMM-based autoencoder (PPNMM-AE)~\cite{R14} and the 3-D-CNN-based autoencoder (3D-NAE) for nonlinear mixtures~\cite{R18}.
The proposed DTU-Net is implemented using PyTorch, and the experiments are conducted on a system equipped with an Intel i9-13900H processor and an Nvidia RTX 4060 GPU.

To ensure a fair comparison between supervised and unsupervised methods, the results of VCA~\cite{R37} are used as endmembers for the supervised PPNMM method. For the other unsupervised methods based on deep learning, the endmembers are either initialized using VCA or follow the procedures described in their respective original papers.

The performance of abundance estimation was assessed by the root mean square error (RMSE) between the estimated and the true abundance tensor, given by 
\begin{equation}
	{\rm RMSE_{abun}}=\sqrt{\frac{1}{NR} \| \boldsymbol{\mathcal{\hat{A}}}  -\boldsymbol{\mathcal{A}} \|_{F}^{2}}.
\end{equation}

The endmember estimation was evaluated by 
the average spectral angle distance (SAD) between the estimated and the true endmember across all endmembers, formulated as
\begin{equation}
	{\rm SAD_{end}}=\frac{1}{R}\sum_{r=1}^{R}\arccos\left(\frac{\langle \hat{\boldsymbol{m}}_{r}, \boldsymbol{m}_{r} \rangle}{\|\hat{\boldsymbol{m}}_{r}\|_{2}  \|\boldsymbol{m}_{r}\|_{2}}\right).
\end{equation}

When evaluating the nonlinear coefficient estimation for PPNMM-generated  datasets, the performance was assessed using the RMSE between the true nonlinear coefficient tensor and the estimated one, defined by
\begin{equation}
	{\rm RMSE_{B}}=\sqrt{\frac{1}{N} \|{\hat{\mathcal{B}}}-\mathcal{B} \|_{F}^{2}}.
\end{equation} 



\subsection{Experiments With Synthetic Datasets}
\subsubsection{Synthetic Datasets Generation}
To comprehensively evaluate the unmixing performance, we generated 12 synthetic datasets, each of size $100 \times 100$ pixels, by varying the number of endmembers, noise levels, and mixing models.

We selected $R=4$ and $R=8$ spectral signatures from the United States Geological Survey (USGS) digital spectral library, each consisting of 224 contiguous bands, to serve as endmembers. The corresponding abundance maps were generated using the Hyperspectral Imagery Synthesis tool for MATLAB~\cite{9439249} \footnote[1]{http://www.ehu.eus/ccwintco/uploads/f/fb/Synthesis.zip}, based on Gaussian random field to ensure piecewise smoothness and incorporate spatial information, thereby resembling realistic abundance distributions.

Using the reference endmembers and abundance fractions, the datasets were synthesized according to three mixing mechanisms: LMM  in~\eqref{LMM}, and the bilinear-based GBM~\cite{R5} and PPNMM~\cite{R7}.
Specifically, the GBM~\cite{R5} is defined by 
\begin{equation}
\boldsymbol{y} = \sum_{r=1}^{R} a_{r}\boldsymbol{m}_{r} + \sum_{i=1}^{R-1} \sum_{j=i+1}^{R} \beta_{ij} a_{i}a_{j} (\boldsymbol{m}_{i} \odot \boldsymbol{m}_{j}) + \boldsymbol{n},
\end{equation}
where $0 \leqslant \beta_{ij} \leqslant 1$ represents the second-order interactions between pairs of distinct endmembers. The values of $\beta_{ij}$ were sampled from a uniform distribution over the interval $[0, 1]$, following~\cite{R5}.
The PPNMM is defined in~\eqref{PPNMM}, where the values of nonlinear coefficient $b$ were uniformly sampled from the interval $[-0.3, 0.3]$, as described in~\cite{R7}.

Finally, Gaussian noise with signal-to-noise ratios (SNRs) of 20 dB and 30 dB was added to each dataset to simulate different noise conditions.

\subsubsection{Hyperparameter Setting}\label{subsec:parameters}
In the experiments on synthetic datasets, the hyperparameters for the proposed DTU-Net were set as follows. Key parameters include the channel dimension $C$ of the embedded feature map in the encoder, the regularization parameter $\alpha$ in the loss function (Eq. \eqref{loss}), and the abundance sparsity parameter $\gamma$ in the Softmax function (Eq. \eqref{softmax}). Additionally, standard neural network hyperparameters such as the number of epochs, learning rate, and weight decay were considered. Notably, a smaller learning rate was applied to the 2-D convolution layer in the decoder to ensure stable training.

\begin{figure}  
	\centering
	\graphicspath{{Pic/}}
	\subfigure[]{
		\label{a_end_1}
		\includegraphics[width=1.5in]{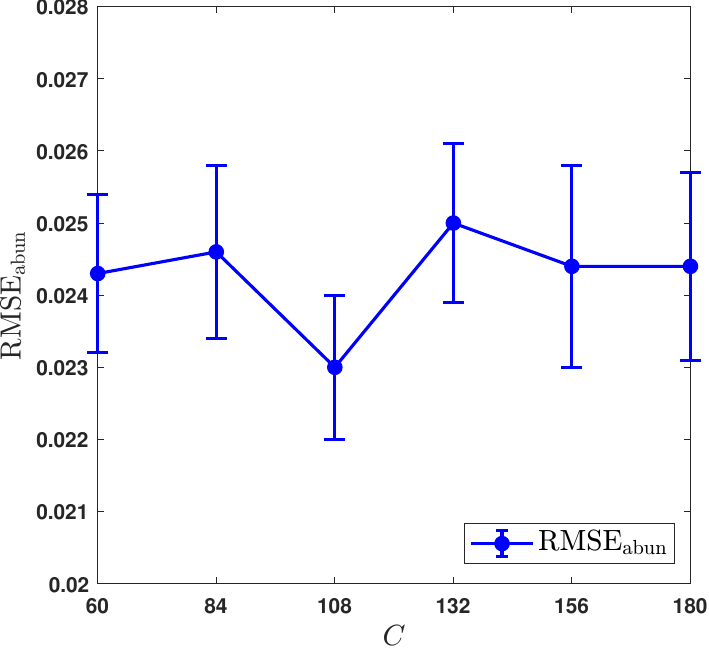} 
	}   
	\subfigure[]{
		\label{b_end_2}
		\includegraphics[width=1.5in]{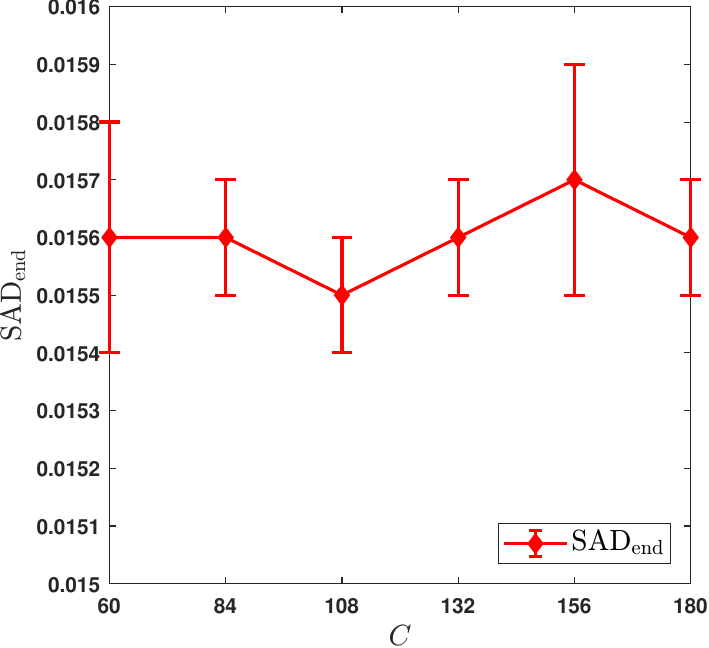}
	}	

	\caption{Variation of metrics with different values of $C$ evaluated on the PPNMM synthetic image with SNR = 30 dB: (a) $\rm RMSE_{abun}$~(b) $\rm SAD_{end}$.}

	\label{F4}
\end{figure}

The choice of $C$ is crucial as it directly correlates with the number of spectral bands in the hyperspectral images. Setting $C$ too low may result in the loss of essential spectral correlations, while an excessively high value can introduce redundant channel information. We recommend that $C$ not exceeds the number of spectral bands, which is 224 in this series of experiments. 
On a PPNMM-generated dataset with  $R = 4$, SNR = 30 dB, a candidate set of values [60, 84, 108, 132, 156, 180] was tested. Figure \ref{F4} illustrates the performance metrics, $\rm RMSE_{abun}$ and $\rm SAD_{end}$, as functions of varying $C$, with ten runs conducted for each value. Based on the results, we set $C = 108$ for all synthetic datasets.

The Adam optimizer was used for training, with the number of epochs fixed at 600 in all the experiments. Other specific hyperparameters for each dataset are summarized in Table \ref{T1}.

%

\begin{table*}[htbp]
\caption{Hyperparameter Setting on Synthetic Datasets }
\setlength{\tabcolsep}{3.6pt}
\renewcommand\arraystretch{0.9}
\centering
	\begin{tabular}{c|c|ccc|ccc}
		\toprule[1pt]
	&	\multirow{2}{*}{\diagbox{{Para.}}{{Data}}} & \multicolumn{3}{c|}{20dB}                                                                     & \multicolumn{3}{c}{30dB}                                                                \\ 
	\cline{3-8} 
	&	& \multicolumn{1}{c|}{LMM}             & \multicolumn{1}{c|}{GBM}             & PPNMM           & \multicolumn{1}{c|}{LMM}           & \multicolumn{1}{c|}{GBM}           & PPNMM         \\
	\cline{1-8} 
\multirow{4}{*}{\rotatebox{270}{$R=4$}}&	$\gamma$  & \multicolumn{1}{c|}{1} & \multicolumn{1}{c|}{1} & 1  & \multicolumn{1}{c|}{1}   & \multicolumn{1}{c|}{1} & 1 \\ 
\cline{2-8}
&$\alpha$ & \multicolumn{1}{c|}{3} & \multicolumn{1}{c|}{1}  & 1& \multicolumn{1}{c|}{1} & \multicolumn{1}{c|}{1}  & $5\times 10^{-1}$          \\ 
\cline{2-8} 
	&	Learning Rate     & \multicolumn{1}{c|}{$6\times 10^{-3}$~/~$10^{-5}$} & \multicolumn{1}{c|}{$6\times 10^{-3}$~/~$10^{-4}$} & $1\times 10^{-2}$~/~$3\times 10^{-5}$ & \multicolumn{1}{c|}{$8\times 10^{-3}$~/~$10^{-5}$} & \multicolumn{1}{c|}{$1\times 10^{-2}$~/~$10^{-5}$} & $1\times 10^{-2}$~/~$10^{-5}$ \\ 
\cline{2-8} 
	&	Weight Decay      & \multicolumn{1}{c|}{$10^{-3}$}       & \multicolumn{1}{c|}{$10^{-3}$}       & $10^{-3}$       & \multicolumn{1}{c|}{$10^{-3}$}     & \multicolumn{1}{c|}{$10^{-3}$}     & $10^{-3}$     \\   
	\hline 
\multirow{4}{*}{\rotatebox{270}{$R=8$}}& $\gamma$ & \multicolumn{1}{c|}{1}  & \multicolumn{1}{c|}{1}  & 1  & \multicolumn{1}{c|}{1} & \multicolumn{1}{c|}{1} & 1  \\ 
\cline{2-8}
& $\alpha$& \multicolumn{1}{c|}{3}& \multicolumn{1}{c|}{1} &1 & \multicolumn{1}{c|}{3}  & \multicolumn{1}{c|}{3}  & 1 \\ 
\cline{2-8} 
& Learning Rate & \multicolumn{1}{c|}{$1\times 10^{-2}$~/~$10^{-5}$} & \multicolumn{1}{c|}{$1\times 10^{-2}$~/~$10^{-4}$} & $1\times 10^{-2}$~/~$10^{-3}$ & \multicolumn{1}{c|}{$8\times 10^{-3}$~/~$10^{-5}$} & \multicolumn{1}{c|}{$1\times 10^{-2}$~/~$10^{-5}$} & $1\times 10^{-2}$~/~$10^{-5}$ \\ 
\cline{2-8}
& Weight Decay      & \multicolumn{1}{c|}{$10^{-3}$}      & \multicolumn{1}{c|}{$10^{-3}$}      & $10^{-4}$      & \multicolumn{1}{c|}{$10^{-3}$}     & \multicolumn{1}{c|}{$10^{-3}$}      & $10^{-3}$      \\ 
\bottomrule[1pt]		
	\end{tabular}
	\label{T1}
\end{table*}

\subsubsection{Results Analysis}
We repeated each experiment ten times and reported the unmixing results as averages in Tables \ref{T2} and \ref{T3}, corresponding to the synthetic datasets with $R=4$ and $R=8$, respectively. Each table contains six datasets, generated by three different mixture models, with two different noise levels.
As observed, across all twelve synthetic datasets, the proposed DTU-Net consistently demonstrates promising unmixing performance in both endmember and abundance estimation, ranking first or second among the compared methods.

\begin{figure*}[!h]
    \centering
    \graphicspath{{Pic/}}
    \subfigure[End\#1]{
        \includegraphics[width=1.3in]{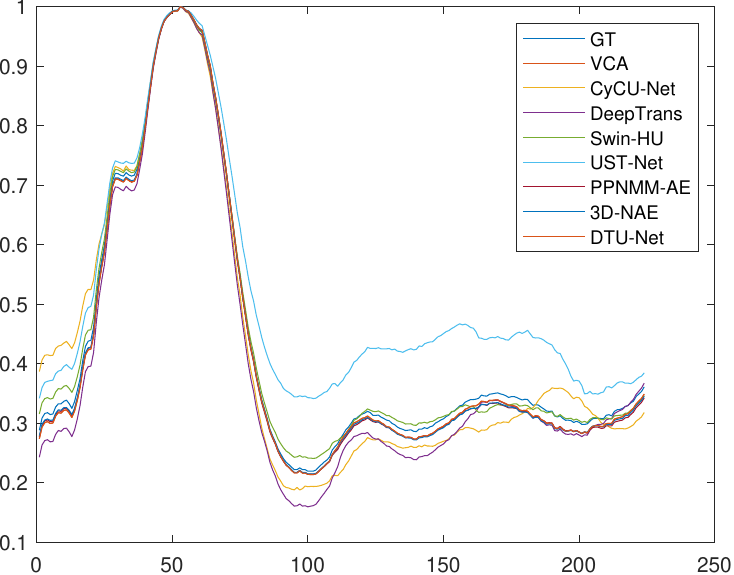}
        \label{a_end_1}
    }    
    \subfigure[End\#2]{
        \includegraphics[width=1.3in]{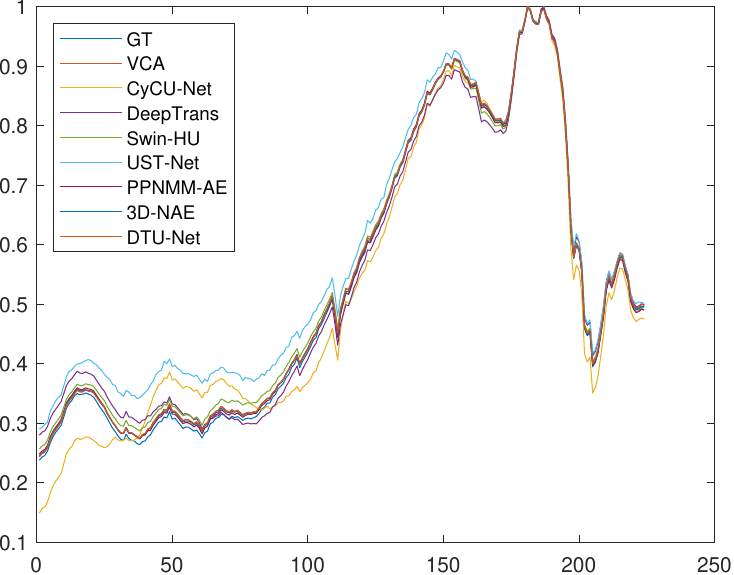}
        \label{b_end_2}
    }
    \subfigure[End\#3]{
        \includegraphics[width=1.3in]{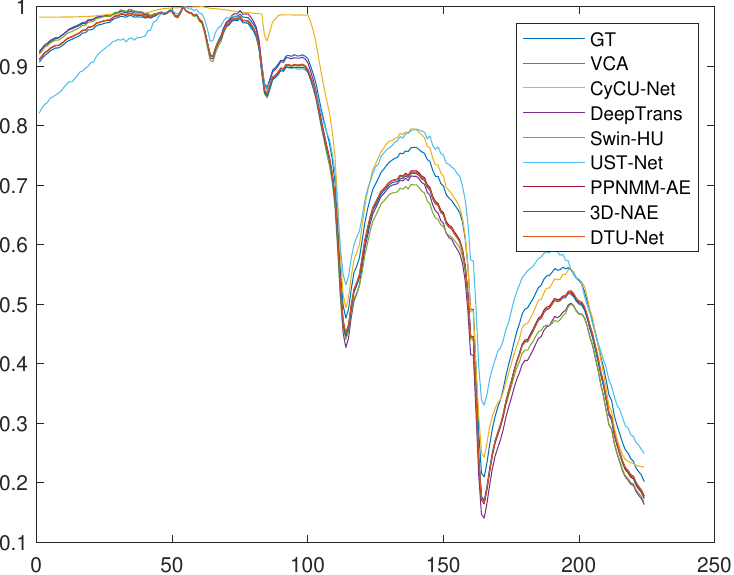}
        \label{c_end_3}
    } 
    \subfigure[End\#4]{
        \includegraphics[width=1.3in]{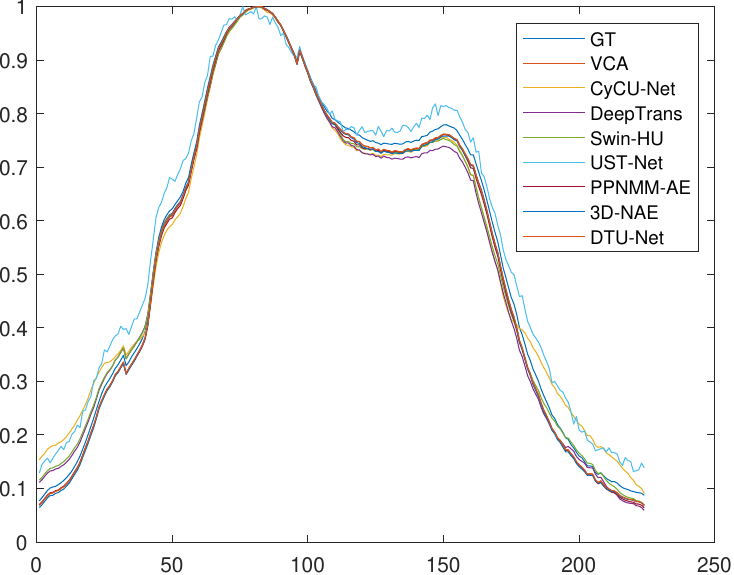}
        \label{d_end_4}
    }       
    \caption{Estimated endmembers by different unmixing methods on a PPNMM-generated dataset with $R = 4$, SNR = 30 dB.}
    \label{F5}
\end{figure*}

%
In terms of endmember extraction, VCA, which operates under the assumption of linear mixing, proves highly effective when the data is relatively clean ($\text{SNR}=30$dB) and the number of endmembers is small ($R=4$), providing the best endmember estimates. This makes VCA a strong initialization method for such scenarios.
For $R=4$ with noise level $\text{SNR}=20$dB, nonlinear unmixing methods, including PPNMM-AE, 3D-NAE, and the proposed DTU-Net, generally offer endmember estimations comparable to or slightly better than VCA, particularly in nonlinear-mixed datasets generated by the GBM and PPNMM models.
As the number of endmembers increases to $R=8$, the proposed DTU-Net outperforms all other methods across the three mixing models and both noise levels, achieving the best endmember extraction. It is followed by VCA and PPNMM-AE, which perform slightly less effectively.
The visualization results for endmember extraction with the synthetic dataset based on PPNMM at $R=4$, $\text{SNR} = 30$dB are shown in Fig. \ref{F5}.

\begin{figure*}[!h]
	\centering
	\graphicspath{{Pic/}}
	\includegraphics[width=7.2in]{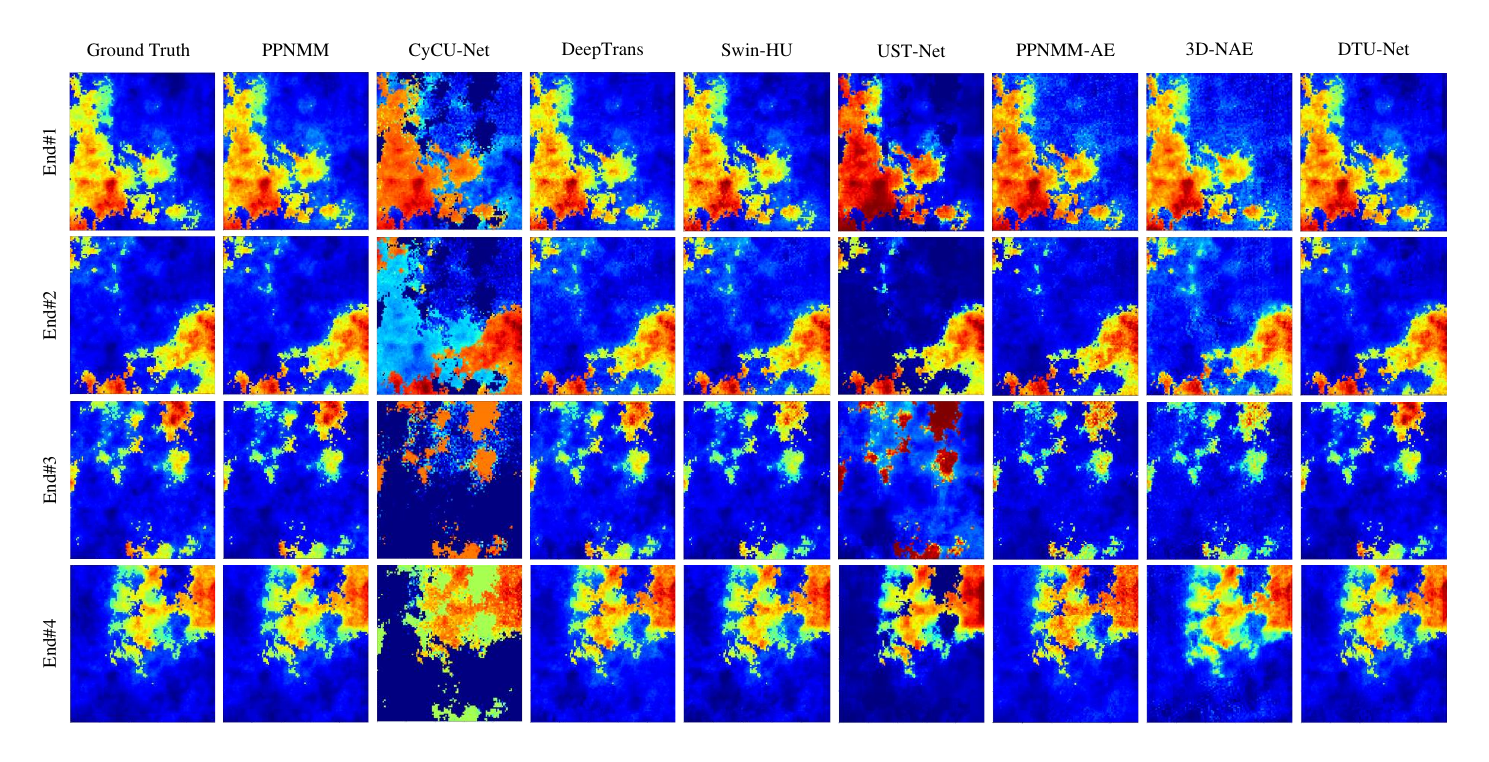}
	\caption{Abundance maps estimation results on a PPNMM-generated dataset with  $R = 4$, SNR = 30 dB. From left to right: Ground Truth, PPNMM, CyCU-Net, DeepTrans, Swin-HU, UST-Net, PPNMM-AE, and the proposed DTU-Net.}
	\centering
	\label{F6}
\end{figure*}

In terms of abundance estimation, the proposed DTU-Net consistently outperforms other deep learning methods based on Transformer and CNN architectures, demonstrating superior accuracy. This performance highlights the effectiveness of the Dilated-Transformer-based encoder, which is capable of capturing multi-scale spatial features and effectively integrating spectral information.
When compared to the traditional PPNMM method, DTU-Net achieves better abundance estimation in nine out of twelve cases, particularly when $R=8$. The PPNMM method, being a supervised approach, relies on prior endmember information. In cases where endmembers can be accurately extracted, especially in scenarios with high SNR or synthetic images generated using LMM, PPNMM tends to superior results.
However, as the number of endmembers increases to $R=8$, DTU-Net, along with other Transformer-based methods such as DeepTrans and Swin-HU, becomes increasingly effective in abundance estimation, especially in noisy cases with $\text{SNR} = 20$ dB. This demonstrates the robustness of the Transformer-based methods, particularly when handling complex and challenging datasets.
The abundance maps for all the comparison methods using the synthetic dataset generated by PPNMM at $R=4$, $\text{SNR} = 30$ dB are visualized in Fig. \ref{F6}.

\begin{figure}[!h]  
	\centering
	\graphicspath{{Pic/}}
	\subfigure[]{
		\label{b_gt}
		\includegraphics[trim=5mm 2mm 10mm 10mm, clip, width=1.4in]{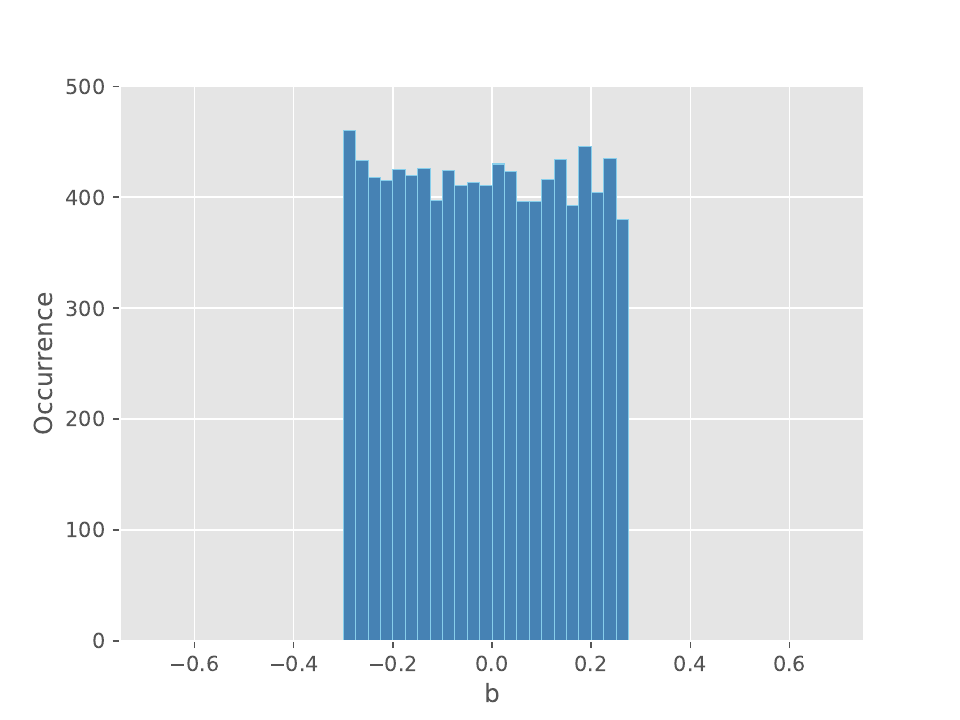}	
		}\hspace{-0.5cm}
	\subfigure[]{
		\label{b_ppnmm}
		\includegraphics[trim=5mm 2mm 10mm 10mm, clip, width=1.4in]{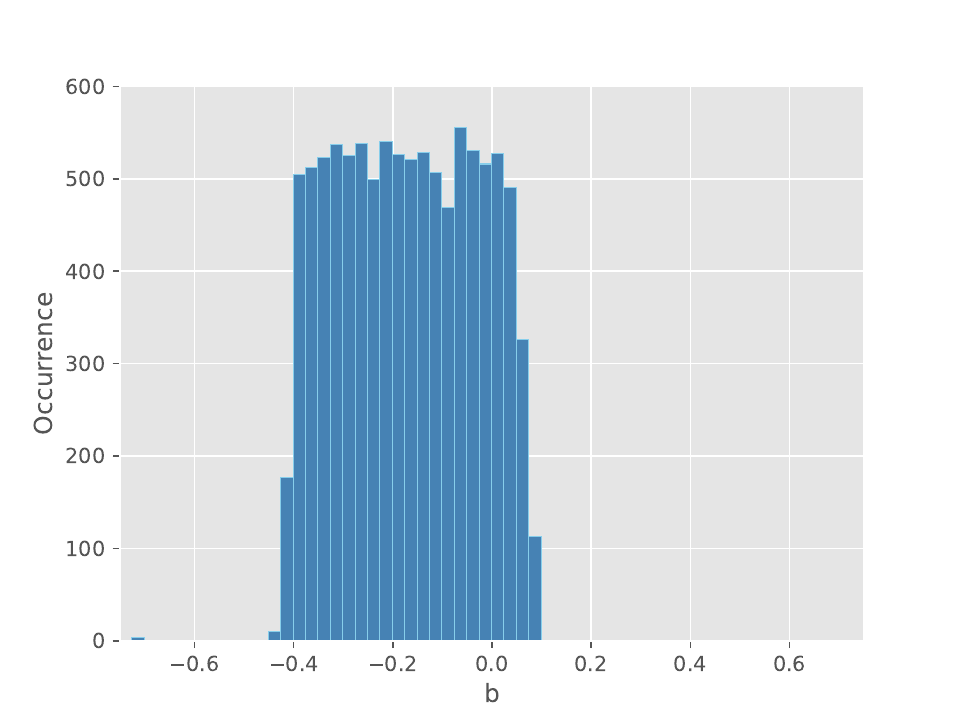}
	}\hspace{0cm}
	\subfigure[]{
		\label{b_ppnmmae}
		\includegraphics[trim=3mm 2mm 10mm 10mm, clip, width=1.4in]{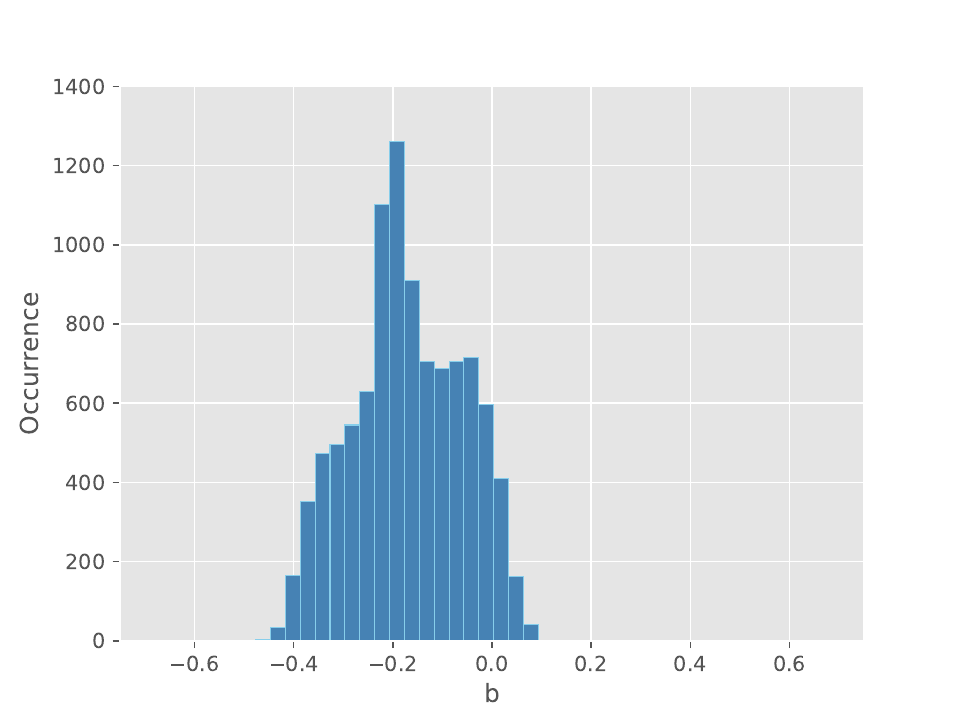}
	} \hspace{-0.5cm}
	\subfigure[]{
		\label{b_DTU-Net}
		\includegraphics[trim=5mm 2mm 10mm 10mm, clip, width=1.4in]{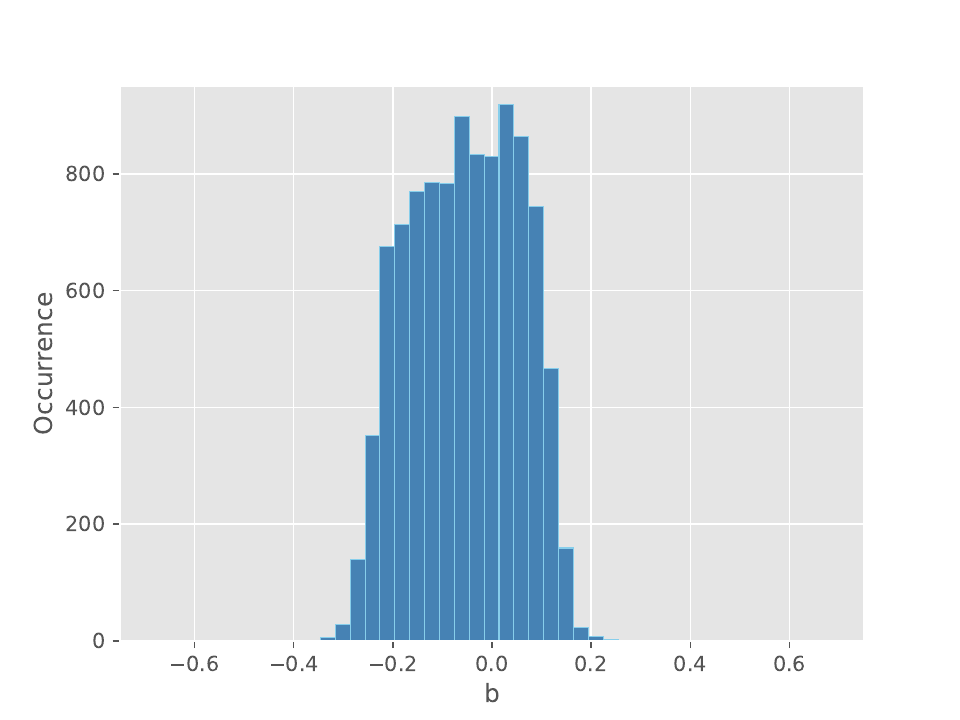}
	}	\hspace{0cm}	
	\caption{Histograms of the nonlinear coefficients $b$ on a PPNMM-generated dataset with  $R = 4$, SNR = 30 dB obtained by: (a) GT (b) PPNMM (c) PPNMM-AE (d) DTU-Net} 
	\label{b_histograms}
\end{figure}

On the four datasets generated according to the PPNMM model, the nonlinear coefficient estimation by the three PPNMM-derived methods, namely PPNMM, PPNMM-AE, and DTU-Net, are quantitatively compared. As observed, DTU-Net, which learns the nonlinear coefficient matrix as a feature map, provides more accurate estimates. 
Fig. \ref{b_histograms} presents the histogram of nonlinear coefficient estimates obtained from three PPNMM-based unmixing methods on a PPNMM-generated dataset with  $R = 4$, SNR = 30 dB. It is observed that the histogram produced by DTU-Net closely aligns with the ground truth (GT) distribution of $b$, while the other two methods exhibit a bias toward negative values. This further demonstrates the effectiveness of DTU-Net in estimating the nonlinear coefficients.

\begin{table*}[htbp]
\caption{Unmixing results on synthetic datasets with $R=4$, SNR=20dB and 30dB, generated using the LMM, GBM, and PPNMM models, averaged over 10 runs. Best and second-best results are marked with circled numbers}
  \renewcommand\arraystretch{1}
	\centering
	\begin{tabular}{cc|rrr|rrr}
		\toprule[1pt]
		\multicolumn{2}{c|}{\multirow{2}{*}{\diagbox{{Unmixing }}{{Mixing}}}}                                  & \multicolumn{3}{c|}{20dB}                                                                                                                        & \multicolumn{3}{c}{30dB}                                                                                                                        \\ \cline{3-8} 
		\multicolumn{2}{c|}{}                                                   & \multicolumn{1}{c|}{LMM}                                 & \multicolumn{1}{c|}{GBM}                                 & \multicolumn{1}{c|}{PPNMM} & \multicolumn{1}{c|}{LMM}                                 & \multicolumn{1}{c|}{GBM}                                 & \multicolumn{1}{c}{PPNMM} \\ \midrule[1pt]
		\multicolumn{1}{c|}{\multirow{8}{*}{$\rm SAD_{end}$}}        & VCA           & \multicolumn{1}{r|}{\textcircled{\tiny2}~0.0065}                   & \multicolumn{1}{r|}{0.0194}                              & \textcircled{\tiny2}~0.0202                     & \multicolumn{1}{r|}{\textcircled{\tiny1}~0.0040}                   & \multicolumn{1}{r|}{\textcircled{\tiny1}~0.0054}                   & \textcircled{\tiny1}~0.0193         \\
		\multicolumn{1}{c|}{}                                   & CyCU-Net      & \multicolumn{1}{r|}{0.0145}                              & \multicolumn{1}{r|}{0.0626}                              & 0.0845                     & \multicolumn{1}{r|}{0.0096}                              & \multicolumn{1}{r|}{0.0660}                              & 0.0631                    \\
		\multicolumn{1}{c|}{}                                   & DeepTrans     & \multicolumn{1}{r|}{0.0159}                              & \multicolumn{1}{r|}{0.0334}                              & 0.0511                     & \multicolumn{1}{r|}{0.0170}                              & \multicolumn{1}{r|}{0.0351}                              & 0.0456                    \\
		\multicolumn{1}{c|}{}                                   & Swin-HU       & \multicolumn{1}{r|}{0.0105}                              & \multicolumn{1}{r|}{0.0346}                              & 0.0410                     & \multicolumn{1}{r|}{0.0118}                              & \multicolumn{1}{r|}{0.0464}                              & 0.0305                    \\
		\multicolumn{1}{c|}{}                                   & UST-Net       & \multicolumn{1}{r|}{0.0827}                              & \multicolumn{1}{r|}{0.0822}                              & 0.0691                     & \multicolumn{1}{r|}{0.0723}                              & \multicolumn{1}{r|}{0.0701}                              & 0.0791                    \\
		\multicolumn{1}{c|}{}                                   & PPNMM-AE      & \multicolumn{1}{r|}{0.0067}                              & \multicolumn{1}{r|}{0.0258}                              & \textcircled{\tiny2}~0.0202          & \multicolumn{1}{r|}{0.0042}                   & \multicolumn{1}{r|}{\textcircled{\tiny2}~0.0055}                   & \textcircled{\tiny2}~0.0202                    \\
		\multicolumn{1}{c|}{}                                   & 3D-NAE        & \multicolumn{1}{r|}{0.0066}                              & \multicolumn{1}{r|}{\textcircled{\tiny2}~0.0191}                   & 0.0274                     & \multicolumn{1}{r|}{0.0131}                              & \multicolumn{1}{r|}{0.0059}                              & 0.0203                    \\
		\multicolumn{1}{c|}{}                                   & Proposed DTU-Net & \multicolumn{1}{r|}{\textcircled{\tiny1}~0.0064}                   & \multicolumn{1}{r|}{\textcircled{\tiny1}~0.0183}                   & \textcircled{\tiny1}~0.0200          & \multicolumn{1}{r|}{\textcircled{\tiny2}~0.0041}                   & \multicolumn{1}{r|}{\textcircled{\tiny1}~0.0054}                   & \textcircled{\tiny1}~0.0193         \\ \hline
		\multicolumn{1}{c|}{\multirow{8}{*}{$\rm RMSE_{abun}$}} & PPNMM         & \multicolumn{1}{r|}{\textcircled{\tiny2}~0.0171}                   & \multicolumn{1}{r|}{\textcircled{\tiny2}~0.0297}                   & \textcircled{\tiny2}~0.0296                     & \multicolumn{1}{r|}{\textcircled{\tiny1}~0.0066}                   & \multicolumn{1}{r|}{\textcircled{\tiny1}~0.0095}                   & \textcircled{\tiny2}~0.0266         \\
		\multicolumn{1}{c|}{}                                   & CyCU-Net      & \multicolumn{1}{r|}{0.1929}                              & \multicolumn{1}{r|}{0.1737}                              & 0.1845                     & \multicolumn{1}{r|}{0.1578}                              & \multicolumn{1}{r|}{0.1649}                              & 0.1683                    \\
		\multicolumn{1}{c|}{}                                   & DeepTrans     & \multicolumn{1}{r|}{0.0239}                              & \multicolumn{1}{r|}{0.0340}                              & 0.0512          & \multicolumn{1}{r|}{0.0191}                              & \multicolumn{1}{r|}{0.0330}                              & 0.0518                    \\
		\multicolumn{1}{c|}{}                                   & Swin-HU       & \multicolumn{1}{r|}{0.0288}                              & \multicolumn{1}{r|}{0.0340}                              & 0.0428                     & \multicolumn{1}{r|}{0.0230}                              & \multicolumn{1}{r|}{0.0344}                              & 0.0421                    \\
		\multicolumn{1}{c|}{}                                   & UST-Net       & \multicolumn{1}{r|}{0.0782}                              & \multicolumn{1}{r|}{0.0850}                              & 0.0692                     & \multicolumn{1}{r|}{0.0866}                              & \multicolumn{1}{r|}{0.0677}                              & 0.0771                    \\
		\multicolumn{1}{c|}{}                                   & PPNMM-AE      & \multicolumn{1}{r|}{0.0215}                              & \multicolumn{1}{r|}{0.0325}                              & 0.0553                     & \multicolumn{1}{r|}{0.0193}                              & \multicolumn{1}{r|}{0.0224}                              & 0.0514                    \\
		\multicolumn{1}{c|}{}                                   & 3D-NAE        & \multicolumn{1}{r|}{0.0649}                              & \multicolumn{1}{r|}{0.0526}                              & 0.0696                     & \multicolumn{1}{r|}{0.0410}                              & \multicolumn{1}{r|}{0.0410}                              & 0.0676                    \\
		\multicolumn{1}{c|}{}                                   & Proposed DTU-Net & \multicolumn{1}{r|}{\textcircled{\tiny1}~0.0144}                   & \multicolumn{1}{r|}{\textcircled{\tiny1}~0.0258}                   & \textcircled{\tiny1}~0.0283          & \multicolumn{1}{r|}{\textcircled{\tiny2}~0.0103}                   & \multicolumn{1}{r|}{\textcircled{\tiny2}~0.0146}                   & \textcircled{\tiny1}~0.0237         \\ \hline
		\multicolumn{1}{c|}{\multirow{3}{*}{$\rm RMSE_{B}$}}    & PPNMM         & \multicolumn{1}{c|}{\multirow{3}{*}{\diagbox{~}{~}}} & \multicolumn{1}{c|}{\multirow{3}{*}{\diagbox{~}{~}}} & 0.1373          & \multicolumn{1}{c|}{\multirow{3}{*}{\diagbox{~}{~}}} & \multicolumn{1}{c|}{\multirow{3}{*}{\diagbox{~}{~}}} & 0.1713         \\
		\multicolumn{1}{c|}{}                                   & PPNMM-AE      & \multicolumn{1}{c|}{}                                    & \multicolumn{1}{c|}{}                                    & \textcircled{\tiny2}~0.1269                     & \multicolumn{1}{c|}{}                                    & \multicolumn{1}{c|}{}                                    & \textcircled{\tiny2}~0.1652                    \\
		\multicolumn{1}{c|}{}                                   & Proposed DTU-Net & \multicolumn{1}{c|}{}                                    & \multicolumn{1}{c|}{}                                    & \textcircled{\tiny1}~0.1007          & \multicolumn{1}{c|}{}                                    & \multicolumn{1}{c|}{}                                    & \textcircled{\tiny1}~0.0934         \\ \bottomrule[1pt]
	\end{tabular}
    \label{T2}
\end{table*}

\begin{table*}[htbp]
\caption{Unmixing results on synthetic datasets with $R=8$, SNR=20dB and 30dB, generated using the LMM, GBM, and PPNMM models, averaged over 10 runs. Best and second-best results are marked with circled numbers}
	\centering
	\begin{tabular}{cc|rrr|rrr}
		\toprule[1pt]
		\multicolumn{2}{c|}{\multirow{2}{*}{\diagbox{{Unmixing}}{{Mixing}}}}                              & \multicolumn{3}{c|}{20dB}                                                                                                                & \multicolumn{3}{c}{30dB}                                                                                                                \\ \cline{3-8} 
		\multicolumn{2}{l|}{}                                               & \multicolumn{1}{c|}{LMM}                             & \multicolumn{1}{c|}{GBM}                             & \multicolumn{1}{c|}{PPNMM} & \multicolumn{1}{c|}{LMM}                             & \multicolumn{1}{c|}{GBM}                             & \multicolumn{1}{c}{PPNMM} \\ \midrule[1pt]
		\multicolumn{1}{c|}{\multirow{8}{*}{$\rm SAD_{end}$}}        & VCA           & \multicolumn{1}{r|}{\textcircled{\tiny2}~0.0232}               & \multicolumn{1}{r|}{\textcircled{\tiny2}~0.0255}               & 0.0384                     & \multicolumn{1}{r|}{\textcircled{\tiny2}~0.0077}               & \multicolumn{1}{r|}{\textcircled{\tiny2}~0.0084}               & \textcircled{\tiny2}~0.0223         \\
		\multicolumn{1}{c|}{}                               & CyCU-Net      & \multicolumn{1}{r|}{0.0257}                          & \multicolumn{1}{r|}{0.0729}                          & 0.0748                     & \multicolumn{1}{r|}{0.0103}                          & \multicolumn{1}{r|}{0.0506}                          & 0.0411                    \\
		\multicolumn{1}{c|}{}                               & DeepTrans     & \multicolumn{1}{r|}{0.0376}                          & \multicolumn{1}{r|}{0.0387}                          & 0.0432                     & \multicolumn{1}{r|}{0.0109}                          & \multicolumn{1}{r|}{0.0125}                          & 0.0400                    \\
		\multicolumn{1}{c|}{}                               & Swin-HU       & \multicolumn{1}{r|}{0.0275}                          & \multicolumn{1}{r|}{0.0290}                          & 0.0364                     & \multicolumn{1}{r|}{0.0161}                          & \multicolumn{1}{r|}{0.0134}                          & 0.0357                    \\
		\multicolumn{1}{c|}{}                               & UST-Net       & \multicolumn{1}{r|}{0.0920}                          & \multicolumn{1}{r|}{0.0947}                          & 0.0847                     & \multicolumn{1}{r|}{0.0908}                          & \multicolumn{1}{r|}{0.0931}                          & 0.0804                    \\
		\multicolumn{1}{c|}{}                               & PPNMM-AE      & \multicolumn{1}{r|}{0.0262}                          & \multicolumn{1}{r|}{0.0289}                          & \textcircled{\tiny2}~0.0355          & \multicolumn{1}{r|}{0.0092}                          & \multicolumn{1}{r|}{0.0107}                          & 0.0234                    \\
		\multicolumn{1}{c|}{}                               & 3D-NAE        & \multicolumn{1}{r|}{0.0393}                          & \multicolumn{1}{r|}{0.0361}                          & 0.0595                     & \multicolumn{1}{r|}{0.0190}                          & \multicolumn{1}{r|}{0.0251}                          & 0.0467                    \\
		\multicolumn{1}{c|}{}                               & Proposed DTU-Net & \multicolumn{1}{r|}{\textcircled{\tiny1}~0.0231}               & \multicolumn{1}{r|}{\textcircled{\tiny1}~0.0247}               & \textcircled{\tiny1}~0.0298          & \multicolumn{1}{r|}{\textcircled{\tiny1}~0.0076}               & \multicolumn{1}{r|}{\textcircled{\tiny1}~0.0083}               & \textcircled{\tiny1}~0.0221         \\ \hline
		\multicolumn{1}{c|}{\multirow{8}{*}{$\rm RMSE_{abun}$}} & PPNMM         & \multicolumn{1}{r|}{0.0349}                          & \multicolumn{1}{r|}{0.0349}                          & 0.0526                     & \multicolumn{1}{r|}{\textcircled{\tiny2}~0.0166}               & \multicolumn{1}{r|}{\textcircled{\tiny1}~0.0173}               & \textcircled{\tiny2}~0.0242         \\
		\multicolumn{1}{c|}{}                               & CyCU-Net      & \multicolumn{1}{r|}{0.1915}                          & \multicolumn{1}{r|}{0.2416}                          & 0.2421                     & \multicolumn{1}{r|}{0.1275}                          & \multicolumn{1}{r|}{0.1412}                          & 0.1496                    \\
		\multicolumn{1}{c|}{}                               & DeepTrans     & \multicolumn{1}{r|}{\textcircled{\tiny2}~0.0321}               & \multicolumn{1}{r|}{0.0342}                          & \textcircled{\tiny2}~0.0422          & \multicolumn{1}{r|}{0.0275}                          & \multicolumn{1}{r|}{0.0439}                          & 0.0353                    \\
		\multicolumn{1}{c|}{}                               & Swin-HU       & \multicolumn{1}{r|}{0.0446}                          & \multicolumn{1}{r|}{\textcircled{\tiny2}~0.0328}               & 0.0423                     & \multicolumn{1}{r|}{0.0210}                          & \multicolumn{1}{r|}{0.0235}                          & 0.0391                    \\
		\multicolumn{1}{c|}{}                               & UST-Net       & \multicolumn{1}{r|}{0.0763}                          & \multicolumn{1}{r|}{0.0886}                          & 0.0751                     & \multicolumn{1}{r|}{0.0748}                          & \multicolumn{1}{r|}{0.0826}                          & 0.0711                    \\
		\multicolumn{1}{c|}{}                               & PPNMM-AE      & \multicolumn{1}{r|}{0.0391}                          & \multicolumn{1}{r|}{0.0449}                          & 0.0662                     & \multicolumn{1}{r|}{0.0355}                          & \multicolumn{1}{r|}{0.0516}                          & 0.0526                    \\
		\multicolumn{1}{c|}{}                               & 3D-NAE        & \multicolumn{1}{r|}{0.0865}                          & \multicolumn{1}{r|}{0.0799}                          & 0.0918                     & \multicolumn{1}{r|}{0.0818}                          & \multicolumn{1}{r|}{0.0767}                          & 0.0862                    \\
		\multicolumn{1}{c|}{}                               & Proposed DTU-Net & \multicolumn{1}{r|}{\textcircled{\tiny1}~0.0252}               & \multicolumn{1}{r|}{\textcircled{\tiny1}~0.0228}               & \textcircled{\tiny1}~0.0328          & \multicolumn{1}{r|}{\textcircled{\tiny1}~0.0164}               & \multicolumn{1}{r|}{\textcircled{\tiny2}~0.0199}               & \textcircled{\tiny1}~0.0229         \\ \hline
		\multicolumn{1}{c|}{\multirow{3}{*}{$\rm RMSE_{B}$}}    & PPNMM         & \multicolumn{1}{c|}{\multirow{3}{*}{\diagbox{~}{~}}} & \multicolumn{1}{c|}{\multirow{3}{*}{\diagbox{~}{~}}} & \textcircled{\tiny2}~0.1429          & \multicolumn{1}{c|}{\multirow{3}{*}{\diagbox{~}{~}}} & \multicolumn{1}{c|}{\multirow{3}{*}{\diagbox{~}{~}}} & 0.1673                    \\
		\multicolumn{1}{c|}{}                               & PPNMM-AE      & \multicolumn{1}{r|}{}                                & \multicolumn{1}{r|}{}                                & 0.1623                     & \multicolumn{1}{r|}{}                                & \multicolumn{1}{r|}{}                                & \textcircled{\tiny2}~0.1577         \\
		\multicolumn{1}{c|}{}                               & Proposed DTU-Net & \multicolumn{1}{r|}{}                                & \multicolumn{1}{r|}{}                                & \textcircled{\tiny1}~0.1042          & \multicolumn{1}{r|}{}                                & \multicolumn{1}{r|}{}                                & \textcircled{\tiny1}~0.1267         \\ \bottomrule[1pt]
	\end{tabular}
	\label{T3}
\end{table*}


\subsection{Experiments With Real Datasets} 
\begin{figure}[htbp]
	\centering
	\graphicspath{{Pic/}}  
	\subfigure[]{
		\label{samson_rgb}
		\includegraphics[width=1in]{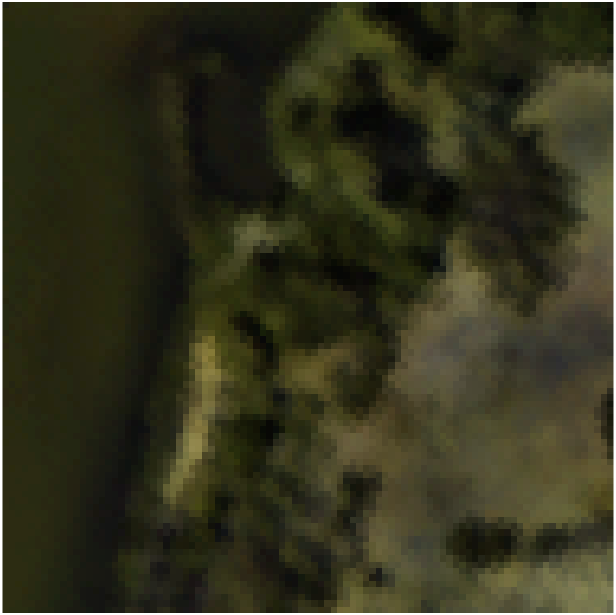}\hspace{-0.22cm}
	}
	\subfigure[]{
		\label{jasper_rgb}
		\includegraphics[width=1in]{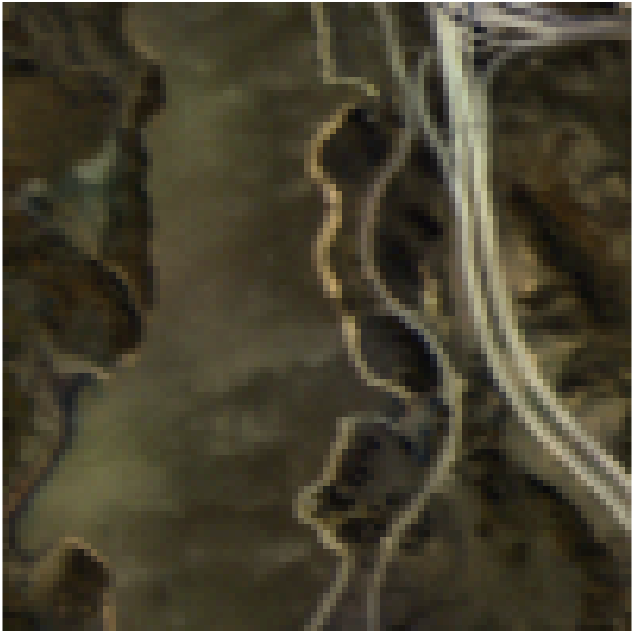}\hspace{-0.22cm}
	}
	\subfigure[]{
		\label{apex_rgb}
		\includegraphics[width=1in]{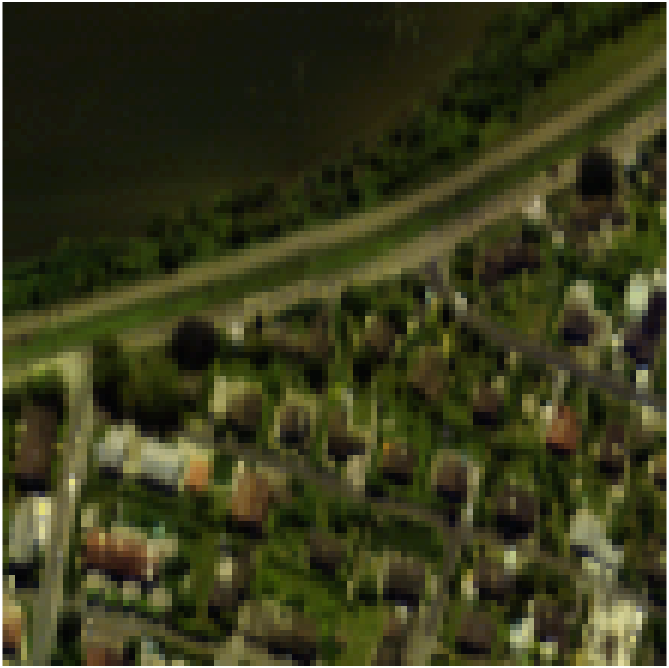}\hspace{-0.22cm}
	}
	\caption{Three hyperspectral images used in the experiments: (a) Samson (b) Jasper Ridge (c) Apex.}
	\label{F7}
\end{figure}

The three real hyperspectral images used in this study are Samson, Jasper Ridge, and Apex, as shown in Fig. \ref{F7}.
It is important to note that, although the GT abundances used to evaluate the unmixing performance of various methods on real datasets are derived under the assumptions of the LMM, with the use of GT endmembers~\cite{R17}\cite{R25}, the proposed DTU-Net remains flexible and capable of handling both linear and nonlinear scenarios.


\subsubsection{Samson}
The Samson dataset was acquired by the SAMSON sensor in Monterey Bay in 2006~\cite{R38}. For the experiment, a sub-image consisting of $95 \times 95$ pixels was selected, spanning 156 spectral channels. These channels cover the wavelength range of 401 nm to 889 nm, with a spectral resolution of 3.13 nm. 
The Samson imagery comprises three primary endmembers: Soil, Tree, and Water. 

The hyperparameters were set as follows: $C=72$, $\gamma=1$, and $\alpha=1$. The learning rate was 0.02 for the linear part of the decoder (i.e., the 2-D convolution layer) and 0.03 for the remaining layers. A weight decay of 0.001 was applied to the entire model and training was conducted over 600 epochs.

\begin{figure*}[h!]
	\centering
	\graphicspath{{Pic/}}
	\includegraphics[width=7in]{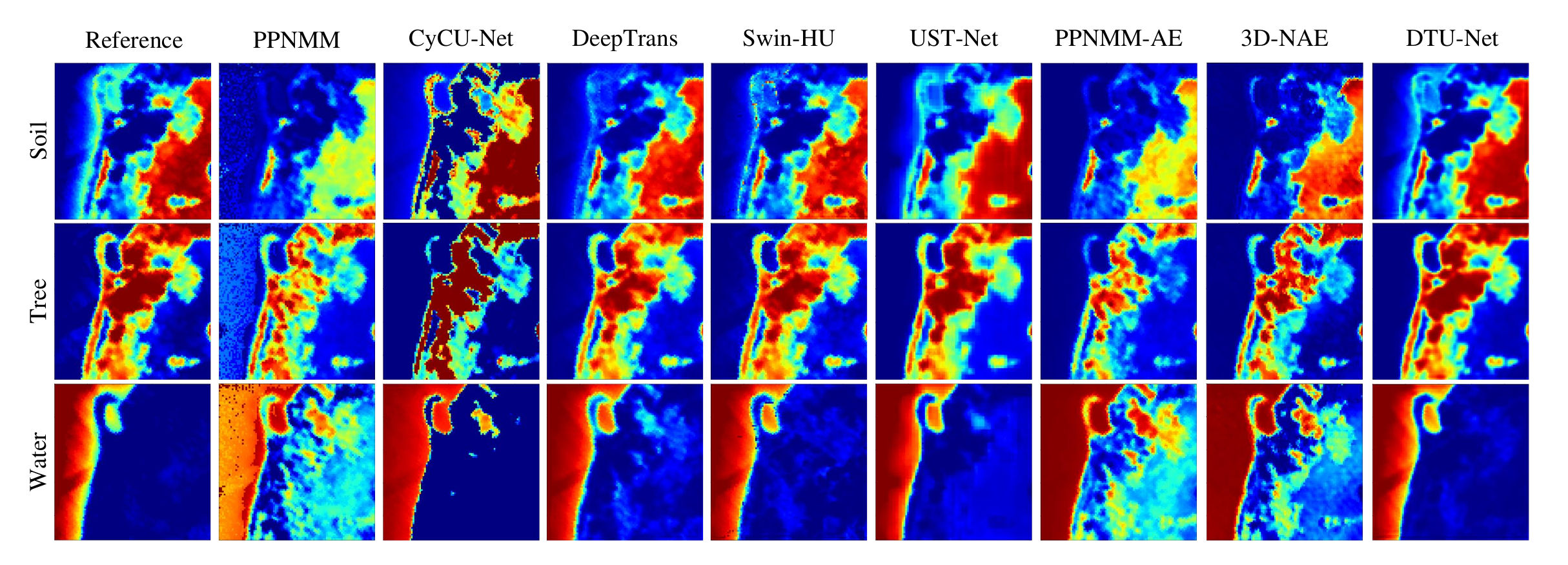}
	\caption{Abundance maps estimation results for the Samson dataset. From left to right: Reference, PPNMM, CyCU-Net, DeepTrans, Swin-HU, UST-Net, PPNMM-AE, and the proposed DTU-Net.}
	\centering
	\label{F8}
\end{figure*}

\begin{figure*}[h!]
	\centering
	\graphicspath{{Pic/}}
	\includegraphics[width=7in]{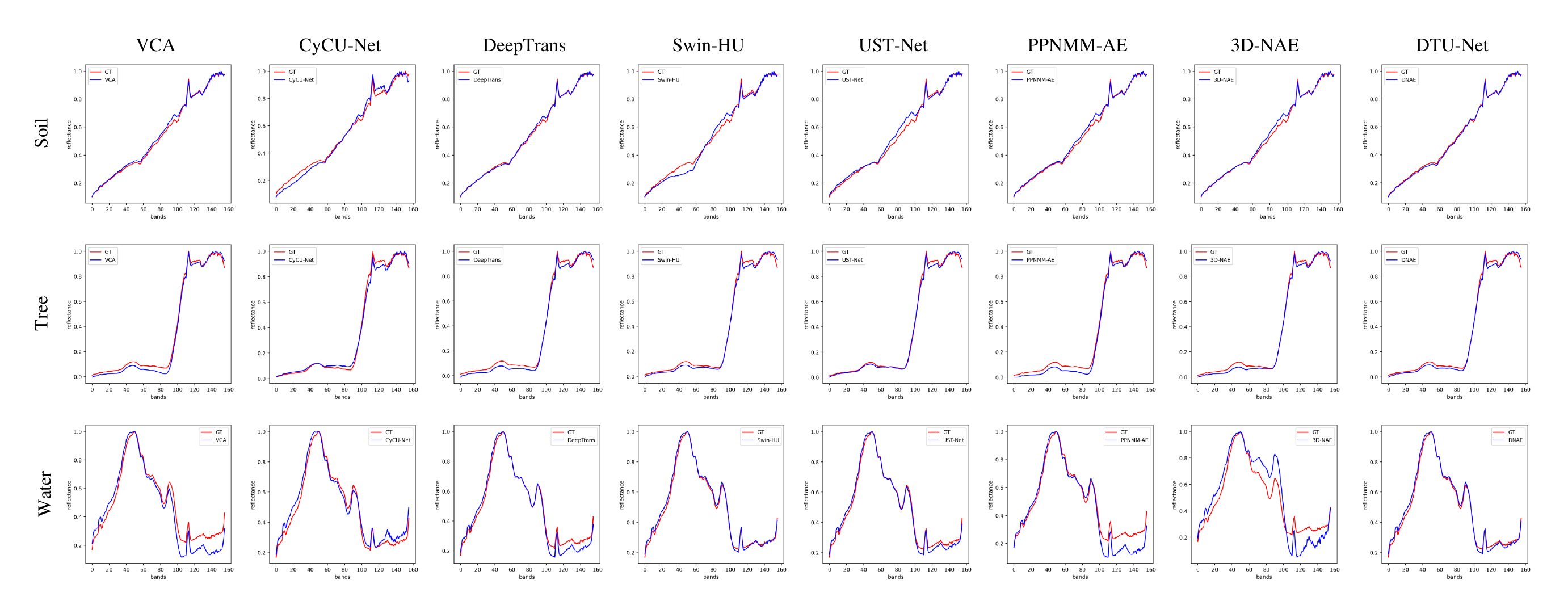}
\caption{Endmember estimation results for the Samson dataset. From left to right: VCA, CyCU-Net, DeepTrans, Swin-HU, UST-Net, PPNMM-AE, and the proposed DTU-Net. GT endmembers in red.}
	\centering
	\label{F9}
\end{figure*}

The unmixing results for the Samson dataset, evaluated by all the compared methods in terms of ${\rm RMSE_{abun}}$ and ${\rm SAD_{end}}$, are reported in Table \ref{T4}. Visualization results for the estimated abundances and extracted endmembers are presented in Fig. \ref{F8} and Fig. \ref{F9}, respectively.
The four Transformer-based methods, namely DeepTrans, Swin-HU, UST-Net and the proposed DTU-Net generally outperform the other comparative methods. Notably, the proposed DTU-Net model demonstrates the best performance for both endmember and abundance estimation.

\begin{table}[htbp]
	\caption{Unmixing Results on Samson Dataset}
	\centering
	\begin{tabular}{c|r|r}
		\toprule[1pt]
		\multicolumn{1}{c|}{\diagbox{{Method.}}{{Metric.}}} & \multicolumn{1}{c|}{$\rm SAD_{end}$} & \multicolumn{1}{c}{$\rm RMSE_{abun}$} \\ \midrule[1pt]
		VCA-PPNMM             & 0.0666                          & 0.2442                                \\
		CyCU-Net              & 0.0503                          & 0.1867                                \\
		DeepTrans             & 0.0421                          & 0.1205                                \\
		Swin-HU               & 0.0374                          & \textcircled{\tiny2}~0.0631                     \\
		UST-Net               & \textcircled{\tiny2}~0.0358               & 0.0853                                \\
		PPNMM-AE              & 0.0678                          & 0.2983                                \\
		3D-NAE                & 0.0753                          & 0.2208                                \\
		Proposed DTU-Net         & \textcircled{\tiny1}~0.0277               & \textcircled{\tiny1}~0.0540                     \\ \bottomrule[1pt]
	\end{tabular}
	\label{T4}
\end{table} 

%


\subsubsection{Jasper Ridge}

The Jasper Ridge dataset was acquired by the Airborne Visible Infrared Imaging Spectrometer (AVIRIS) over Jasper Ridge in central California, USA~\cite{R39}. The original image consists of $512 \times 614$ pixels, covering the spectral range from 380 to 2500 nm across 224 channels, with a spectral resolution of 9.46 nm. For the experiment, a sub-image of $100 \times 100$ pixels was selected for analysis. To account for atmospheric effects and water vapor absorption, channels 1-3, 108-112, and 154-166 were excluded, resulting in a reduced dataset of 198 spectral bands. This image primarily contains four endmembers: Tree, Water, Soil, and Road.

In this experiment, we set $C=96$, $\gamma=1$, and $\alpha=5.5$. The learning rate was set to 0.0086, and a weight decay of $10^{-4}$ was applied to all layers of the network. Training was conducted over 600 epochs.

\begin{figure*}[h!]
	\centering
	\graphicspath{{Pic/}}
	\includegraphics[width=7in]{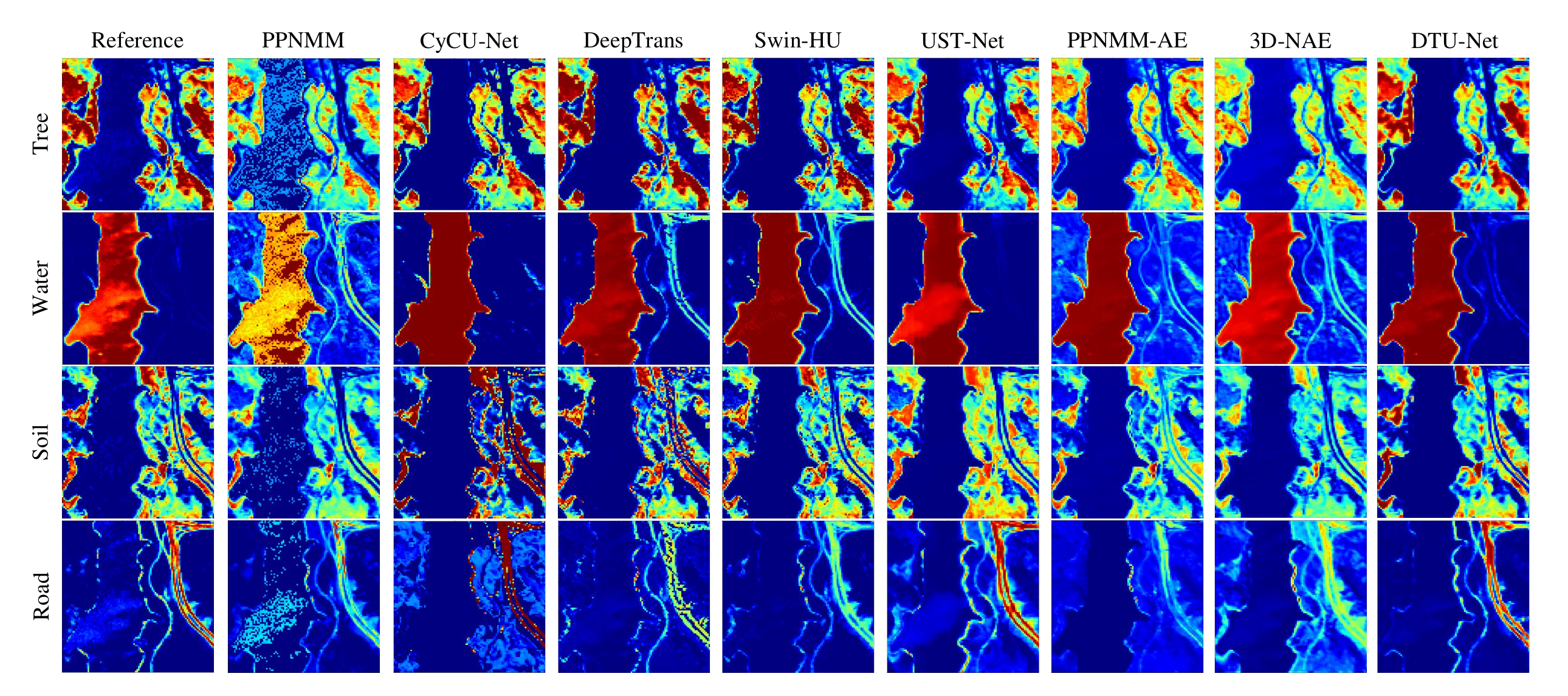}
	\caption{Abundance maps estimation results for the Jasper Ridge dataset. From left to right: Reference, PPNMM, CyCU-Net, DeepTrans, Swin-HU, UST-Net, PPNMM-AE, and the proposed DTU-Net.}
	\centering
	\label{F10}
\end{figure*}

\begin{figure*}[h!]
	\centering
	\graphicspath{{Pic/}}
	\includegraphics[width=7in]{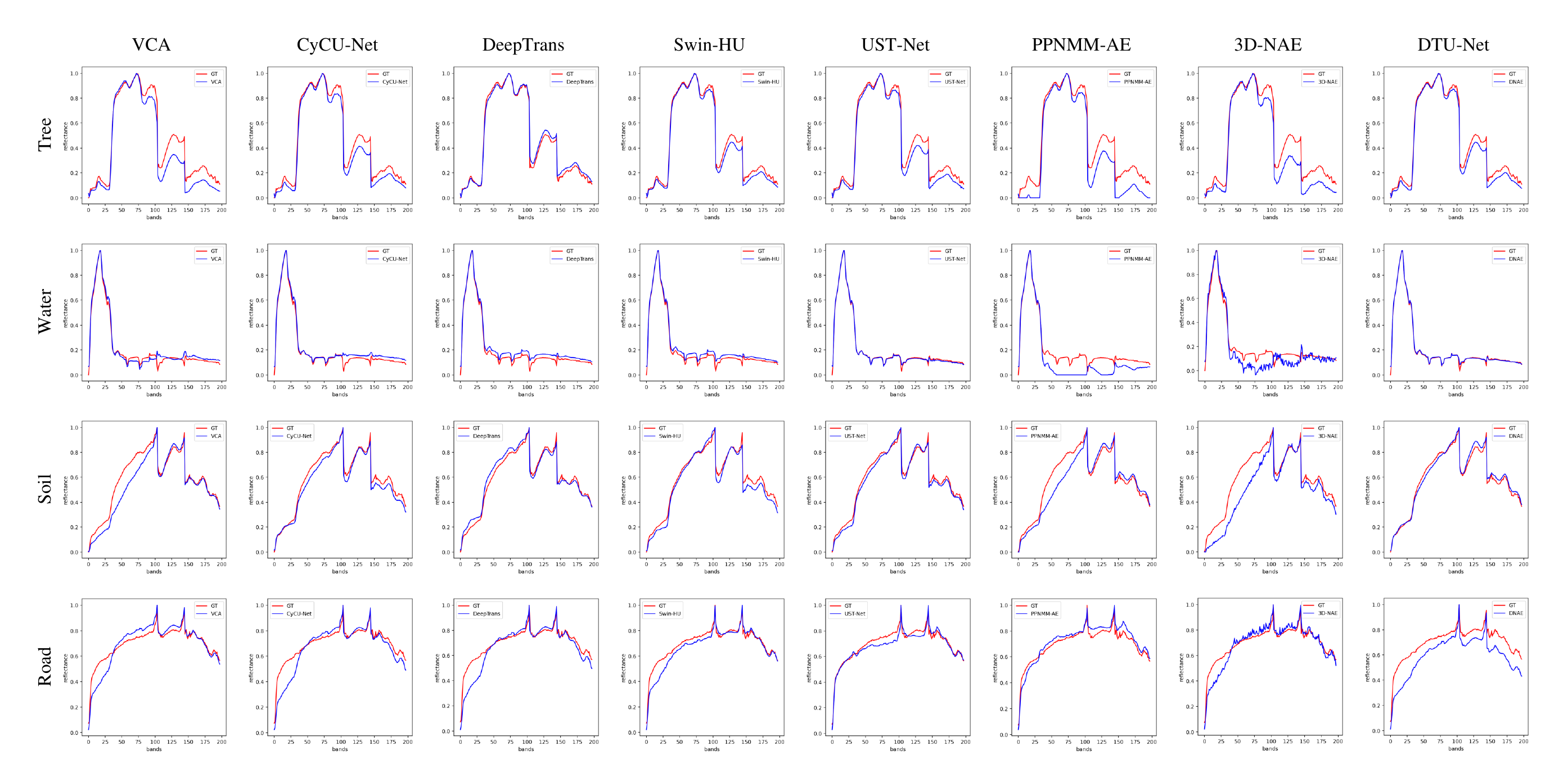}
	\caption{Endmembers estimation results for the Jasper Ridge dataset. From left to right: VCA, CyCU-Net, DeepTrans, Swin-HU, UST-Net, PPNMM-AE, and the proposed DTU-Net. GT endmembers in red.}
	\centering
	\label{F11}
\end{figure*}

Table \ref{T5} presents a quantitative comparison of unmixing performance on Jasper Ridge, while the comparative visualization results for the estimated abundances and extracted endmembers are shown in Figs. \ref{F10} and \ref{F11}, respectively.
Compared to Samson dataset, the Jasper Ridge dataset comprises a greater number of endmembers and exhibits a more complex spatial feature distribution. 
The Transformer-based approaches demonstrate superior performance on this dataset, compared to CNN-based and traditional unmixing techniques.
Notably, the proposed DTU-Net yielded the most accurate abundance estimates and the second-best endmember estimates, following UST-Net.

\begin{table}[htbp]
	\caption{Unmixing Results on Jasper Ridge}
	\centering
	\begin{tabular}{c|r|r}
		\toprule[1pt]
		\multicolumn{1}{c|}{\diagbox{{Method}}{{Metric}}} & \multicolumn{1}{c|}{$\rm SAD_{end}$} & \multicolumn{1}{c}{$\rm RMSE_{abun}$} \\ \midrule[1pt]
		VCA-PPNMM             & 0.1110                          & 0.1623                                \\
		CyCU-Net              & 0.0780                          & 0.1305                                \\
		DeepTrans             & 0.0699                          & 0.1078                                \\
		Swin-HU               & 0.0618                          & 
		0.0963                                \\
		UST-Net               &\textcircled{\tiny1}~0.0565               & \textcircled{\tiny2}~0.0889                                \\
		PPNMM-AE              & 0.1650                          & 0.1594                                \\
		3D-NAE                & 0.1565                          & 0.1751                                \\
		Proposed DTU-Net         & \textcircled{\tiny2}~0.0584               & \textcircled{\tiny1}~0.0773                     \\ \bottomrule[1pt]
	\end{tabular}
	\label{T5}
\end{table}


\subsubsection{Apex}
The Apex dataset was acquired by the airborne imaging spectrometer, Airborne Prism Experiment (APEX), during an APEX flight campaign in June 2011~\cite{R40}. The image used in the experiment is a cropped portion of $110 \times 110$ pixels, containing 285 spectral bands from the original dataset, which spans spectral channels from 413 nm to 2420 nm. This dataset includes four endmembers: Water, Tree, Road, and Roof.
%

For the experiments on the Apex dataset, the parameters were set as follows: $C=216$, $\gamma=1.2$, and $\alpha=30$. Due to the low accuracy of the initial endmembers extracted by the VCA method, a higher learning rate of 0.01 and a lower weight decay of $10^{-8}$ were applied to the linear component of the decoder. For the remaining layers of the model, the learning rate was set to 0.0016, and the weight decay was set to $3 \times 10^{-4}$. Training was conducted over 600 epochs.

\begin{figure*}[h!]
	\centering
	\graphicspath{{Pic/}}
	\includegraphics[width=7in]{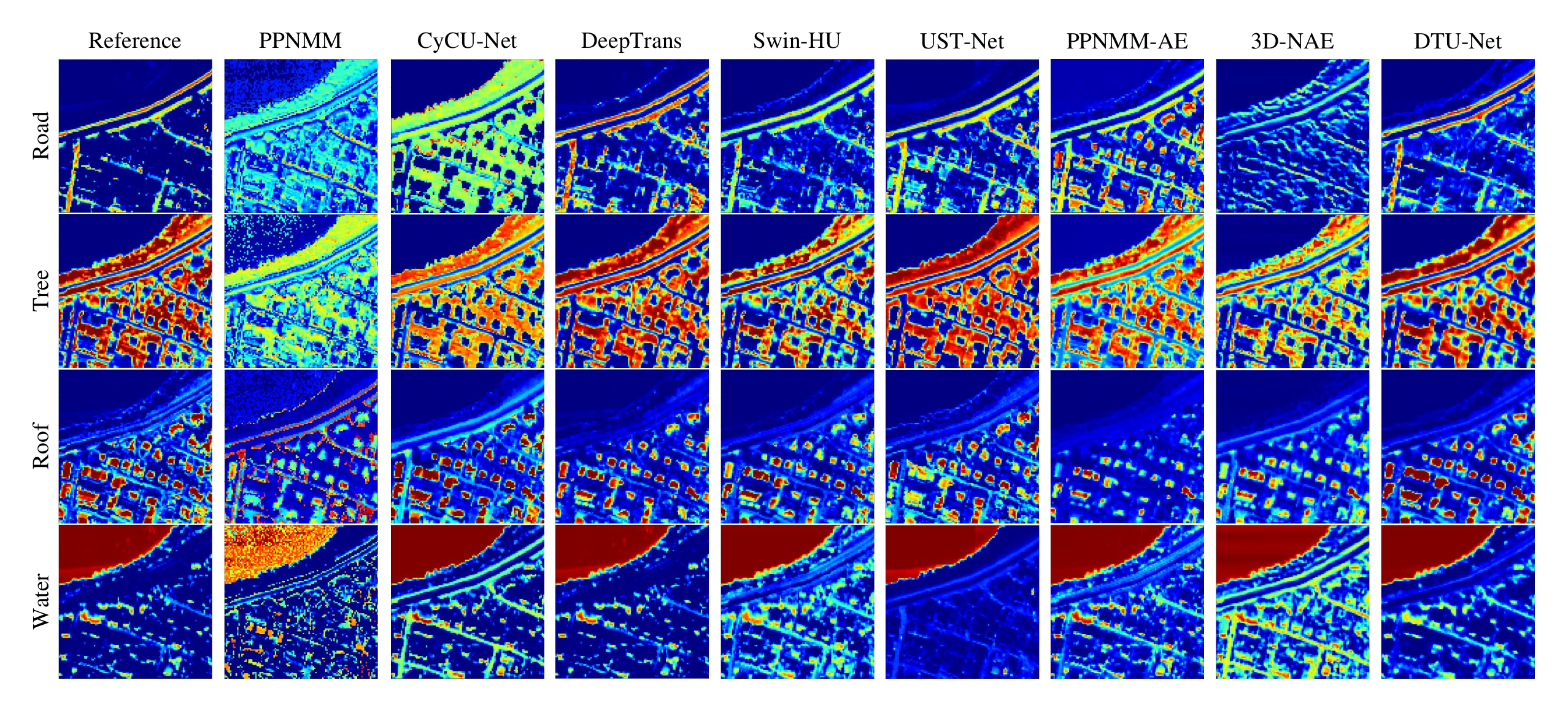}
	\caption{Abundance maps estimation results for the Apex dataset. From left to right: Reference, PPNMM, CyCU-Net, DeepTrans, Swin-HU, UST-Net, PPNMM-AE, and the proposed DTU-Net.}
	\centering
	\label{F12}
\end{figure*}

\begin{figure*}[h!]
	\centering
	\graphicspath{{Pic/}}
	\includegraphics[width=7in]{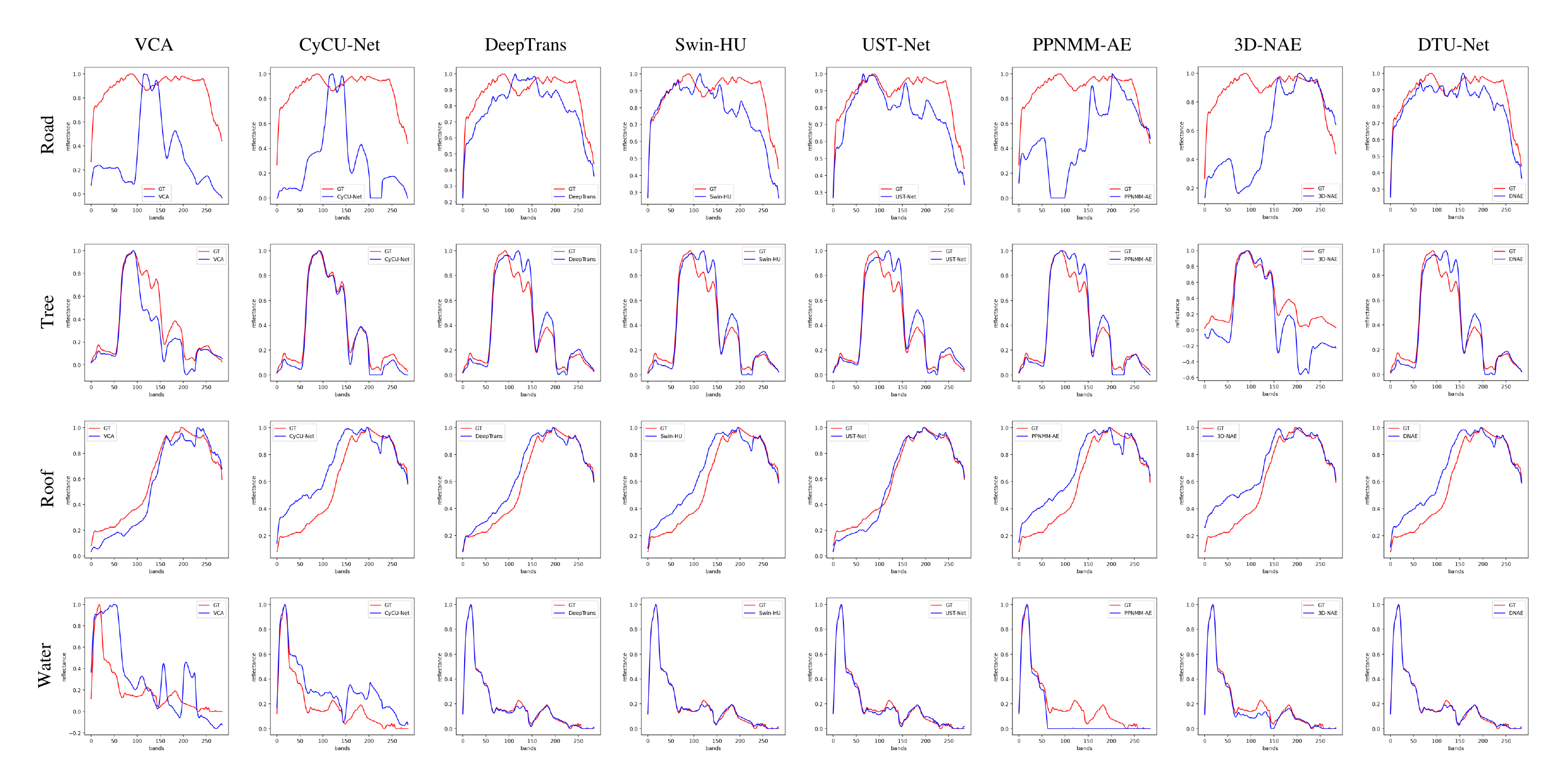}
	\caption{Endmembers estimation results for the Apex dataset. From left to right: VCA, CyCU-Net, DeepTrans, Swin-HU, UST-Net, PPNMM-AE, and the proposed DTU-Net. GT endmembers in red.}
	\centering
	\label{F13}
\end{figure*}

A quantitative comparison of unmixing performance across different methods on the Apex dataset is presented in Table \ref{T6}, with the visualization results for the estimated abundances and extracted endmembers shown in Fig. \ref{F12} and Fig. \ref{F13}, respectively. The Apex dataset contains richer spectral information with more spectral bands and exhibits more complex spatial characteristics, making it a greater challenge for unmixing tasks. As shown in Table \ref{T6}, the proposed DTU-Net achieves the best unmixing performance in terms of both endmember and abundance estimation among all the compared methods, followed by DeepTrans in endmember extraction and Swin-HU in abundance estimation, demonstrating the effectiveness of DTU-Net in handling complex datasets. Fig. \ref{F12} and Fig. \ref{F13} reveal that while the estimation of the endmember ``road" and its corresponding abundance proves challenging for most methods, Transformer-based approaches achieve good results. Notably, the proposed model provides the most accurate extraction of this endmember and its abundance, with results that most closely align with the GT.

\begin{table}[htbp]
	\caption{Unmixing Results on Apex}
	\centering
	\begin{tabular}{c|r|r}
		\toprule[1pt]
		\multicolumn{1}{c|}{\diagbox{{Method}}{{Metric}}} & \multicolumn{1}{c|}{$\rm SAD_{end}$} & \multicolumn{1}{c}{$\rm RMSE_{abun}$} \\ \midrule[1pt]
		VCA-PPNMM             & 0.3958                          & 0.3179                                \\
		CyCU-Net              & 0.2768                          & 0.1779                                \\
		DeepTrans             & \textcircled{\tiny2}~0.0953                          & 0.1351                                \\
		Swin-HU               & 0.1156                          & 
		\textcircled{\tiny2}~0.1279                                \\
		UST-Net               & 0.0957               & 0.1529                                \\
		PPNMM-AE              & 0.2890                          & 0.1629                                \\
		3D-NAE                & 0.2958                          & 0.2247                                \\
		Proposed DTU-Net         & \textcircled{\tiny1}~0.0952               & \textcircled{\tiny1}~0.0978                     \\ \bottomrule[1pt]
	\end{tabular}
	\label{T6}
\end{table}


\subsection{Ablation Experiments}
The encoder of the proposed DTU-Net incorporates a dual-branch structure designed to extract multi-scale spatial and spectral features from the input image, respectively. To validate the contribution of each branch, we conducted an ablation study on three real datasets. The results of this ablation study, presented in Table \ref{T7}, demonstrate that the collaborative use of both branches, extracting and fusing spatial and spectral features, outperforms the individual performance of each branch in terms of unmixing accuracy. When analyzing the separate performance of each branch, it is evident that the multi-scale spatial feature extraction branch yields superior results when operating independently, highlighting the effectiveness of the Dilated Transformer in capturing multi-scale spatial dependencies in hyperspectral images. 
\begin{table*}[]
\caption{Ablation Experiment Results on Samson, Jasper Ridge and Apex}
\setlength{\tabcolsep}{6pt}
\centering
\begin{tabular}{c|c|c|c|c|c|c|c}
\toprule
\multicolumn{2}{c|}{} & \multicolumn{2}{c|}{Samson} & \multicolumn{2}{c|}{Jasper Rigde} & \multicolumn{2}{c}{Apex} \\
\hline
$\mathcal{F}_\text{spatial}$ & $\mathcal{F}_\text{spectral}$ & $\rm RMSE_{abun}$ & $\rm SAD_{end}$ & $\rm RMSE_{abun}$ & $\rm SAD_{end}$ & $\rm RMSE_{abun}$ & $\rm SAD_{end}$ \\
\hline
\checkmark & $\times$ & 0.0728 & 0.0956 & 0.0727 & 0.0933 & 0.1098 & 0.1944 \\
$\times$ & \checkmark & 0.1211 & 0.3908 & 0.4071 & 0.2316 & 0.2727 & 0.3253 \\
\checkmark & \checkmark & \textbf{0.0277} & \textbf{0.0540} & \textbf{0.0584} & \textbf{0.0773} & \textbf{0.0952} & \textbf{0.0978} \\
\bottomrule
\end{tabular}
\label{T7}
\end{table*}


\section{Conclusion}\label{Sec:V}
This paper presented DTU-Net, an end-to-end multi-scale Dilated Transformer-based nonlinear unmixing network, built upon the autoencoder structure. The encoder has two branches: the first is a Dilated Transformer branch that captures multi-scale spatial dependencies using Sliding Window Dilated Attention (SWDA) and Multi-Scale Dilated Attention (MSDA) to represent long-range and multi-scale spatial correlations. The second branch incorporates 3D-CNNs and channel attention mechanisms to model spectral dependencies.
The fused feature of the two branches is transformed to abundances.
The decoder is designed according on the PPNMM, first mimicking the linear mixture part and then extending it with the PPNMM mechanism. By explicitly modeling the relationships between endmembers, abundances, and nonlinear coefficients, the decoder captures both linear and nonlinear mixing characteristics. Additionally, by learning the abundances and nonlinear coefficients as pixel-wise features, the decoder provides both flexibility and interpretability, enabling accurate and robust nonlinear unmixing across diverse datasets.
The effectiveness of the proposed DTU-Net was proved on both synthetic and real datasets, by comparing with the PPNMM-derived and other unmixing networks.
It is noteworthy that compared to the abundance estimation with advanced deep learning techniques such as Transformers, current unmixing networks for endmember estimation primarily rely on the weight matrix of fully connected layers. Future work will aim to address this challenge by developing more robust and adaptive techniques for endmember estimation in unmixing tasks.

\bibliographystyle{IEEEtran}
\bibliography{DNAE}

\end{document}